\documentclass[11pt]{article}

\usepackage[preprint]{acl}

\usepackage{times}
\usepackage{latexsym}

\usepackage[T1]{fontenc}

\usepackage[utf8]{inputenc}

\usepackage{microtype}

\usepackage{inconsolata}

\usepackage{graphicx}

\usepackage{tikz}
\usepackage{amssymb}
\usepackage{booktabs}
\usepackage{makecell}
\usepackage{fontawesome5}
\usetikzlibrary{positioning,arrows.meta,calc,fit,backgrounds,shapes.geometric}

\usepackage{needspace}
\usepackage{algorithm}
\usepackage[noend]{algpseudocode}
\usepackage{stfloats} %
\newcommand{\qualrule}{{\color{black!35}\hrule height 0.6pt}}
\newenvironment{qualexample}[1]{%
  \par\Needspace{5\baselineskip}%
  \vspace{6pt}\qualrule\vspace{4pt}%
  \setlength{\parskip}{1.5pt}\setlength{\parindent}{0pt}%
  \noindent\textbf{#1}\par\nobreak}{%
  \par\vspace{4pt}\qualrule\vspace{6pt}}

\title{Agentic Clustering: Controllable Text Taxonomies via Multi-Agent Refinement}

\author{
 \textbf{Simon Löwe\textsuperscript{1},
 Emily Silcock\textsuperscript{2}}
\\
 \textsuperscript{1}Burning Glass Institute,
 \textsuperscript{2}Harvard University
\\
 \small{
   \textbf{Correspondence:} \href{mailto:emilysilcock@fas.harvard.edu}{emilysilcock@fas.harvard.edu}
 }
}

\begin{document}
\maketitle
\begin{abstract}
Recent text-clustering methods use large language models to propose a cluster taxonomy from a corpus and then assign each text to it. These pipelines are fundamentally programmatic: the sequence of LLM calls and the rules for stopping, merging, and splitting clusters are fixed in code in advance, so they generalise poorly across corpora of different structure and cannot easily incorporate user-supplied constraints such as a target cluster count or a clustering intent. We propose an agentic alternative in which an orchestrator LLM inspects the state of the discovery process at each step and dispatches one of a small set of specialised agents --- proposer, synthesizer, auditor, investigator, and critic --- adapting the pipeline to the corpus rather than executing a fixed one. On seven public text-clustering benchmarks the method achieves state-of-the-art performance, beating the strongest prior baseline by 47\% in mean ARI.
\end{abstract}

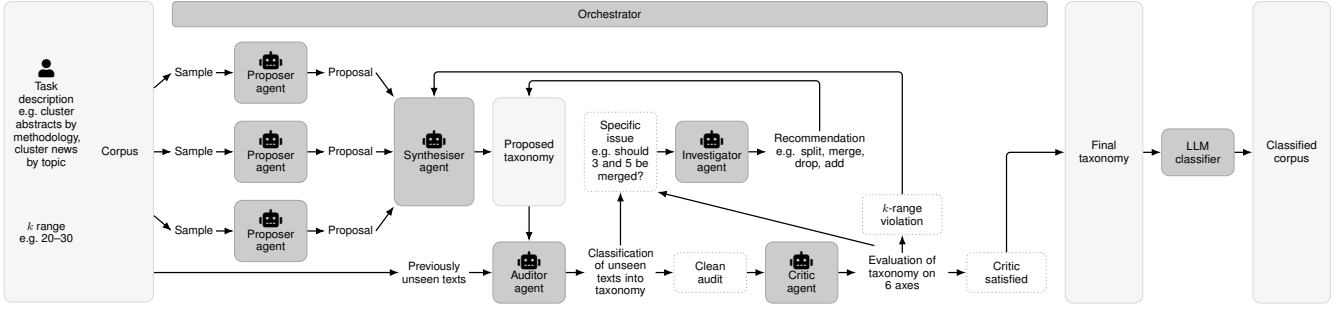
\begin{figure*}[t]
  \centering
  \makebox[\textwidth][c]{%
    \resizebox{1.1\textwidth}{!}{%
        \begin{tikzpicture}[
  font=\sffamily\small,
  node distance=10mm and 14mm,
  >=Latex,
  rounded corners=2pt,
  box/.style={
    rectangle,
    rounded corners=6pt,
    draw=gray!45,
    line width=0.8pt,
    fill=white,
    align=center,
    minimum height=9mm,
    text width=30mm,
    inner sep=3pt
  },
  actor/.style={
    box,
    text width=18.5mm,
    minimum height=17mm,
    fill=gray!40,
    draw=gray!70
  },
  yellowbox/.style={
    box,
    text width=24mm,
    fill=gray!7,
    draw=gray!30
  },
  dottedbox/.style={
    rectangle,
    rounded corners=4pt,
    draw=gray!70,
    fill=white,
    line width=0.8pt,
    dotted,
    align=center,
    minimum height=11mm,
    inner xsep=2.5mm,
    inner ysep=3pt
  },
  inputsub/.style={
    dottedbox,
    dashed,
    fill=gray!7,
    inner sep=3pt
  },
  textonly/.style={
    align=center,
    inner sep=2pt,
    font=\sffamily\small
  },
  arrow/.style={
    -{Stealth[length=2.3mm, width=1.7mm, inset=0.5mm, round]},
    line width=0.75pt,
    draw=black!75,
    rounded corners=4pt,
    shorten >=0.5pt
  },
  faintarrow/.style={arrow}
]

\node[draw=gray!30, fill=gray!7, rounded corners=6pt, line width=0.8pt,
      minimum width=28mm, minimum height=83mm, inner sep=0pt] (input) {};

\node[anchor=north] (usericon) at ([yshift=-3mm]input.north) {\Large\faUser};

\node[inputsub, text width=22mm, below=2mm of usericon] (task) {%
  Task\\ description\\
  e.g.\ cluster abstracts by methodology, cluster news by topic};

\node[inputsub, text width=22mm, below=4mm of task] (kin) {%
  $k$~range\\
  e.g.\ 20--30};

\node[textonly, right=5mm of input, yshift=22mm] (sample1) {Sample};
\node[textonly, right=5mm of input]              (sample2) {Sample};
\node[textonly, right=5mm of input, yshift=-22mm] (sample3) {Sample};

\node[actor, text width=16mm, right=5mm of sample1] (prop1) {{\Large\faRobot}\\[2.5pt] Proposer\\ agent};
\node[actor, text width=16mm, right=5mm of sample2] (prop2) {{\Large\faRobot}\\[2.5pt] Proposer\\ agent};
\node[actor, text width=16mm, right=5mm of sample3] (prop3) {{\Large\faRobot}\\[2.5pt] Proposer\\ agent};

\node[textonly, right=5mm of prop1] (proposal1) {Proposal};
\node[textonly, right=5mm of prop2] (proposal2) {Proposal};
\node[textonly, right=5mm of prop3] (proposal3) {Proposal};

\node[actor, right=5mm of proposal2, text width=20mm, minimum height=30mm] (synth)
  {{\Large\faRobot}\\[2.5pt] Synthesizer\\ agent};

\node[yellowbox, right=5mm of synth, text width=16mm, minimum height=30mm] (taxonomy) {
  Proposed\\ taxonomy
};

\node[dottedbox, minimum height=22mm, right=5mm of taxonomy] (issue) {%
  Specific\\ issue\\
  e.g.\ should\\ 3 and 5 be\\ merged?%
};

\node[actor, right=5mm of issue] (investigator) {{\Large\faRobot}\\[2.5pt] Investigator\\ agent};

\node[textonly, right=5mm of investigator] (recommendation) {
  Recommendation\\
  e.g. split, merge,\\
  remove, add
};

\node[actor, text width=16mm, below=9.77mm of taxonomy] (auditor) {{\Large\faRobot}\\[2.5pt] Auditor\\ agent};

\node[textonly] (classification) at (issue |- auditor.center)
  {Classification\\ of unseen texts\\ into taxonomy};

\node[textonly] (unseen) at (synth |- auditor.center)
  {Previously\\ unseen texts};

\node[inputsub, text width=22mm, minimum height=9mm] (corpus) at (input.center |- auditor.center)
  {Corpus};

\node[dottedbox, right=5mm of classification] (cleanaudit) {
  Audit\\ complete
};

\node[actor, text width=16mm, right=5mm of cleanaudit] (critic) {{\Large\faRobot}\\[2.5pt] Critic\\ agent};

\node[textonly, right=5mm of critic] (evaluation) {
  Evaluation of\\
  taxonomy on\\
  6 axes
};

\node[dottedbox, right=5mm of evaluation] (satisfied) {
  Critic\\ satisfied
};

\coordinate (channel) at ($(issue.south)!0.6!(issue.south |- critic.north)$);
\node[dottedbox, minimum height=9mm, anchor=south] (krange) at ([xshift=7mm, yshift=2mm]evaluation.north |- channel) {
  $k$-range\\
  violation
};

\path let \p1 = (input.north west), \p2 = (input.south west), \p3 = (satisfied.east) in
  coordinate (colnw) at (\x3 + 5mm, \y1)
  coordinate (colsw) at (\x3 + 5mm, \y2)
  coordinate (colsize) at (21mm, {(\y1 - \y2 - 14mm) / 3});

\path let \p1 = (colsize) in
  node[yellowbox, text width=18mm, minimum width=\x1, minimum height=\y1,
       anchor=north west, inner sep=0pt] (final) at (colnw) {Final\\ taxonomy}
  node[actor, text width=18mm, minimum width=\x1, minimum height=\y1,
       below=7mm of final, inner sep=0pt] (classifier) {LLM\\ classifier}
  node[yellowbox, text width=18mm, minimum width=\x1, minimum height=\y1,
       anchor=south west, inner sep=0pt] (classified) at (colsw) {Classified\\ corpus};

\draw[arrow] (input) -- (sample1.west);
\draw[arrow] (input) -- (sample2);
\draw[arrow] (input) -- (sample3.west);

\draw[arrow] (sample1) -- (prop1);
\draw[arrow] (sample2) -- (prop2);
\draw[arrow] (sample3) -- (prop3);

\draw[arrow] (prop1) -- (proposal1);
\draw[arrow] (prop2) -- (proposal2);
\draw[arrow] (prop3) -- (proposal3);

\draw[arrow] (proposal1.east) -- (synth);
\draw[arrow] (proposal2.east) -- (synth);
\draw[arrow] (proposal3.east) -- (synth);

\draw[arrow] (synth) -- (taxonomy);
\draw[arrow] (taxonomy) -- (auditor);

\draw[faintarrow]
  (corpus.east) -- (unseen.west);

\draw[faintarrow] (unseen.east) -- (auditor.west);
\draw[arrow] (auditor.east) -- (classification.west);

\draw[arrow] (classification) -- (cleanaudit);
\draw[arrow] (cleanaudit) -- (critic);
\draw[arrow] (critic) -- (evaluation);
\draw[arrow] (evaluation) -- (satisfied);

\draw[arrow] (classification.north) -- (issue.south);
\draw[arrow] ([xshift=-6mm]evaluation.north) -- ([xshift=-6mm]evaluation.north |- channel)
  -| ([xshift=6mm]issue.south);

\draw[arrow] (issue) -- (investigator);
\draw[arrow] (investigator) -- (recommendation);

\draw[arrow]
  (recommendation.north) -- ++(0, 14mm) -| (taxonomy.north);

\draw[arrow] (evaluation.north -| krange.south) -- (krange.south);

\draw[arrow] (satisfied.north) |- (final.west);

\draw[arrow] (final.south) -- (classifier.north);
\draw[arrow] (classifier.south) -- (classified.north);

\draw[arrow]
  (krange.north) -- ++(0, 34mm) -| (synth.north);

\path let \p1 = ([xshift=5mm]input.north east), \p2 = ([xshift=-5mm]final.north west) in
  node[draw=gray!70, rounded corners=3pt, fill=gray!40, line width=0.8pt,
       font=\sffamily\small, anchor=north west,
       minimum height=7mm, minimum width={\x2 - \x1},
       inner xsep=2pt, inner ysep=0pt]
  at (\p1) {Orchestrator};
  \end{tikzpicture}%
    }%
  }
  \caption{Architecture of the agentic-clustering pipeline.} %
  \label{fig:architecture}
\end{figure*}

\section{Introduction}

A common task in exploring large unstructured text corpora is clustering the texts into meaningful groups. Clustering approaches could be applied for generating themes from customer feedback, discovering emergent trends in job advertisements, or identifying common themes in scientific literature. Further, a text corpus does not always have a single, obvious clustering, and may have many such groups. For example, a corpus of academic abstracts could be grouped by topic, methodology, or dataset. We refer to these as clustering `intents'.

Large language models (LLMs) have led to improvements across many NLP tasks. However, the easy application of LLMs to text clustering is limited by the fact that a whole dataset cannot generally fit into the context window of any LLM. %

Recent work has focused on using LLMs for clustering by splitting the task into two separate sub-tasks. First, an LLM is fed samples of the dataset, usually in chunks, and generates a taxonomy of cluster names. Second, texts are assigned to a label from this taxonomy, using the taxonomy as the LLM prompt --- a standard classification task \citep{pham-etal-2024-topicgpt, wang-etal-2023-prompttopic, huang-he-2024-text-clustering}. The most recent of these methods add further passes that revise, merge or split clusters once an initial draft has been formed \citep{wan2024tntllm}. However, every method in this line is fundamentally \emph{programmatic}: the sequence of LLM calls, the rules for when to stop, merge or split, and the kinds of evidence each step is allowed to consult are all fixed in code in advance. As a consequence, these pipelines are tuned to the datasets they were designed on and do not generalise well to corpora with different structure --- in particular, they are inflexible in the granularity of the clusters they produce, performing worse than dense embedding approaches when faced with a dataset of a different granularity.

We propose instead an \emph{agentic} approach to text clustering. Rather than executing a fixed pipeline, an orchestrator agent inspects the state of the discovery process at each step and dispatches one of a small set of specialised agents: a \emph{proposer} that drafts candidate clusters from samples of the corpus, a \emph{synthesizer} that merges those drafts into a single taxonomy, an \emph{auditor} that probes the taxonomy against previously unseen texts, an \emph{investigator} that resolves specific concerns by issuing targeted retrieval queries, and a \emph{critic} that evaluates the taxonomy along several axes. The pipeline is no longer hardcoded but assembled on the fly.

This agentic framing matters because clustering is a problem with no single right answer. The right number of clusters, the right level of abstraction, and the right way to resolve a boundary case all depend on the corpus, and a programmatic pipeline must commit to a fixed answer for each of these in advance. An agentic pipeline can instead interrogate the corpus, observe what the data actually supports, and steer itself accordingly; the same agent capabilities apply unchanged to a 77-class banking-intent corpus and an 18-class spoken-domain corpus, with the orchestrator adapting its dispatches to the structure each one exhibits. The same flexibility lets the orchestrator accept user constraints --- a target $k$-range and an optional free-text task description of the clustering lens --- and forward them to every agent without rewriting the pipeline. 

We evaluate our approach on seven public text-clustering benchmarks, and find that it achieves state-of-the-art performance, exceeding the strongest prior baseline by 47\% in mean ARI.

We release this approach as a Claude Code plugin, for easy deployment, under the MIT License at \url{https://github.com/emilysilcock/agentic-clustering}.\footnote{The released artifact ships only the agent prompts and orchestration code, with no data derived from the research-licensed benchmarks used for evaluation.}

\section{Related work}
\label{sec:related}

\paragraph{Probabilistic topic models.}
Traditional probabilistic topic models are still widely used for text clustering, particularly in the social sciences and humanities. This line of work treats each document as a mixture over latent topics and each topic as a distribution over words, recovering both from a bag-of-words representation via variational EM or Gibbs sampling. Latent Dirichlet Allocation \citep{blei2003lda} is the canonical example.\footnote{See \citet{jagarlamudi-etal-2012-incorporating} and \citet{akash-etal-2024-topicadapt} for extensions.} These approaches are limited by relying on a bag-of-words document representation, and by the need to pre-specify the number of clusters.

\paragraph{Dense embeddings with classical clustering.}
The advent of transformers \citep{vaswani2017attention} led to a new paradigm for text clustering, where a transformer is used to encode the documents into a dense embedding vector, and then classical clustering algorithms (e.g. $k$-means, HDBSCAN, etc.) are used to cluster the vectors.
Dense embeddings generally outperform probabilistic topic models. Sentence-BERT \citep{reimers-gurevych-2019-sentence-bert} fine-tunes BERT so that semantically similar sentences land nearby in vector space, and $k$-means on top yields semantic clusters at a pre-specified $k$. BERTopic \citep{grootendorst2022bertopic} adds UMAP dimensionality reduction and HDBSCAN, allowing variable $k$; Top2Vec \citep{angelov2020top2vec} jointly learns document and word vectors in the same UMAP+HDBSCAN regime. %

\paragraph{LLM-improved dense embeddings.}
Two lines of work use LLMs to improve dense-embedding clustering. The first shapes the embedding space itself: ClusterLLM \citep{zhang-etal-2023-clusterllm} fine-tunes the encoder from LLM-answered triplet queries and uses pairwise prompts to set hierarchical granularity; \citet{viswanathan2024fewshot} apply LLMs to enrich inputs, supply pairwise constraints, and post-correct assignments; IDAS \citep{deraedt-etal-2023-idas} clusters utterances jointly with LLM-written labels; and LLMEdgeRefine \citep{feng-etal-2024-llmedgerefine} re-assigns only low-confidence boundary points. The second replaces the encoder with an LLM embedder: \citet{petukhova2024llm} show this beats BERT-based baselines.

\paragraph{LLM-driven taxonomy generation.}

A further line of work has focused on using LLMs to generate the taxonomy itself. PromptTopic \citep{wang-etal-2023-prompttopic} runs every text through an LLM to generate a candidate topic, and then uses an LLM to merge all the candidate topics into a final taxonomy. \citet{huang-he-2024-text-clustering} follows a similar approach, but passes the corpus to the LLM in batches of 15 texts at a time. They then use a lightweight classifier to assign each text to a topic. Passing all texts through an LLM may be infeasible for a large corpus, so TopicGPT \citep{pham-etal-2024-topicgpt} passes the texts one at a time, but asks it to compare to a growing taxonomy of topics, with early stopping if 200 texts are processed with no addition to the taxonomy. They then also follow this with a merging step, and a lightweight classifier.

Building on these approaches, the most recent line of work has added further refinement stages into taxonomy generation. TnT-LLM \citep{wan2024tntllm} uses an LLM to summarize a sample of documents, and splits this sample into minibatches. They use the first mini-batch to initialise a taxonomy, and then subsequent minibatches to refine the taxonomy, adding, splitting or merging clusters. Additionally, unlike the previous methods, TnT-LLM allows for user-specified cluster intents.

\section{Method}
\label{sec:method}

Our method replaces the fixed pipeline of prior LLM-driven taxonomy work with a set of specialised agents coordinated by an orchestrator that dispatches them in a state-dependent order. The agent capabilities themselves recall earlier work --- the proposer samples corpus batches as in \citet{huang-he-2024-text-clustering}, the synthesizer merges drafts as in \citet{wang-etal-2023-prompttopic}, and the auditor, investigator and critic refine the output along the lines of \citet{wan2024tntllm} --- but the orchestrator decides which to dispatch, when, and what to do with each return value, removing the hardcoded stop, merge and split rules that constrain those prior pipelines. The framework also accepts user-specified granularity and clustering intent as inputs that the orchestrator forwards to every agent.

\subsection{Inputs}
\label{sec:method-inputs}

The method takes as input an unlabelled corpus $D$ of texts together with two user constraints. The first is a target \emph{$k$-range} $[k_{\min}, k_{\max}]$ on the number of clusters, which the synthesizer respects when merging proposals and which the critic checks at review time. The second is an optional free-text \emph{task description} stating the lens through which clusters should be formed (e.g.\ ``cluster by issue type'' or ``focus on actionable categories for a support team''). When supplied, the task description is forwarded verbatim to every agent and acts as the primary constraint when resolving ambiguous cases; when omitted, agents discover clusters from the data alone. The two constraints operate on orthogonal axes --- the task description shapes \emph{what} counts as a cluster, the $k$-range shapes \emph{how many}.

\subsection{Orchestrator}
\label{sec:method-orchestrator}

The discovery loop is driven by an \emph{orchestrator} that, at each step, inspects the current workspace state and dispatches one of the agents described in the subsections below. The orchestrator does not itself read the corpus or make structural changes to the taxonomy, though it may make minor refinements. It also owns the decision of what to do with each agent's return value --- whether to apply an investigator's recommendation, how to route a critic's finding, or when to declare the taxonomy final --- and is the only component that persists across the entire run. Algorithm~\ref{alg:orchestrator} (Appendix~\ref{sec:appendix-orchestrator}) gives pseudocode for the method.

\subsection{Proposer}
\label{sec:method-proposer}

The proposer is the agent responsible for generating candidate cluster hypotheses directly from the corpus. Each proposer reads a sample of texts and emits a draft cluster set --- a list of named clusters with short natural-language descriptions, together with the identifiers of the texts it took as evidence for each cluster and the identifiers of any texts it could not place. Multiple proposers are dispatched in parallel so that the downstream synthesis stage can identify which structure is robust across independent looks at the corpus rather than relying on any single draw. The number of proposers is determined dynamically by the orchestrator: it begins with six to seven from different stylistic angles, with additional proposers dispatched when the requested $k$-range is wide.

Proposers receive uniform random samples of $D$, and the orchestrator decides whether these samples are disjoint or shared between proposers; drawn ids are added to a run-wide seen set and excluded from all later draws. The size of each sample is chosen by the orchestrator at dispatch time and scales inversely with the typical text length --- short utterances are sampled in batches of several hundred while longer documents are sampled in batches of tens --- so that each proposer sees enough corpus to identify recurring structure while keeping the sample within its working context.

All proposers share the same base prompt and user task description, but each also receives a \emph{style directive}: a short instruction, written by the orchestrator at dispatch time, that sets the granularity or framing at which the proposer looks for clusters. In our runs these ranged from granularity targets (e.g.\ splitting fine intents, or merging when in doubt) to alternative framings (e.g.\ grouping by action or by object). A style directive cannot override the user task description or the $k$-range. Varying it across proposers surfaces structure at several granularities at once and lets the synthesizer arbitrate, rather than committing to a single granularity in advance.

\subsection{Synthesizer}
\label{sec:method-synthesizer}

Once the proposer phase has produced a set of independent draft cluster sets, the synthesizer is dispatched to merge them into a single proposed taxonomy. The synthesizer reads every proposer's output in full, identifies which clusters across proposers refer to the same underlying concept, and emits a proposed taxonomy in which each cluster is named, described, and supported by evidence from the contributing proposer outputs.

The synthesizer relies on two signals to decide which proposed clusters to merge. The first is overlap in the natural-language content of the proposals --- clusters with similar names and descriptions across proposers are candidates for merging. The second is direct inspection of the underlying texts: the synthesizer fetches a handful of evidence texts from each candidate cluster by id and reads them to confirm that the two definitions actually describe the same kind of text. The proposed taxonomy is constrained to lie within the user-specified $k$-range $[k_{\min}, k_{\max}]$; the synthesizer's merge decisions are biased toward fewer or more clusters as needed to fall inside this range. Clusters that appear in only one proposer's output are retained but flagged as lower-confidence, since they have only been observed in a single draw of the corpus.

Cluster names and descriptions are written under a set of \emph{description quality} rules: descriptions must define the concept that a cluster captures rather than summarise the observed evidence texts; they may not cite specific text identifiers, statistics, or numeric figures from the corpus; and any boundary clarification must be phrased as a generalisable principle rather than an instance-specific correction. The intent is that a reader unfamiliar with the corpus could use the description to assign new texts correctly without seeing the evidence.

\subsection{Auditor}
\label{sec:method-auditor}

The auditor validates the proposed taxonomy against texts that no agent has yet seen: it pulls a new sample of texts and tests, text by text, whether each can be assigned to one of the proposed clusters.

The auditor draws only from texts outside the run-wide seen set. Its samples are of two kinds: a uniform random draw, from which coverage is estimated, and a stratified draw that targets clusters with fewer than five audit assignments by retrieving texts similar to their descriptions; the taxonomy cannot be finalised until every cluster reaches this floor. The auditor therefore sees only texts that contributed neither to the proposers' draft cluster sets nor to the synthesizer's merge decisions, so any cluster that holds up under audit holds up on evidence outside its own derivation.

For each text in the sample the auditor produces a single assignment: the id of the best-fitting cluster in the proposed taxonomy together with an integer \emph{confidence score} from 1 (forced guess) to 5 (obvious fit). When no cluster is a defensible fit, the auditor instead assigns the special label \emph{none}; these assignments are recorded without a confidence score and treated separately during aggregation.

The auditor's output combines a per-text assignment table with a set of aggregate statistics: the fraction of texts assigned versus left unclustered (the \emph{coverage}), mean confidence overall and per cluster, and a list of clusters whose mean assigned confidence falls below a threshold of 3 on this 1--5 scale. Clusters on this list are flagged as \emph{weak} and become the natural targets for follow-up by the agents introduced in subsequent subsections.

\subsection{Investigator}
\label{sec:method-investigator}

When the auditor or critic flags an issue, the investigator is dispatched. The investigator targets a specific concern (e.g.\ ``should c3 and c5 be merged?'') and decides whether the taxonomy should be modified in response.

To gather the texts relevant to its question, the investigator issues query-driven retrieval calls against the corpus: TF-IDF similarity against a free-text query when the concern is conceptual, the texts that recent audits assigned to a particular cluster when the concern is internal coherence, or specific text identifiers when verifying a candidate merge or split. It reads the retrieved texts, weighs them against the proposed taxonomy and the user task description, and forms a structured recommendation.

Recommendations take one of six forms: \emph{merge} two or more clusters into one, \emph{split} a cluster into several, \emph{rename} a cluster (and rewrite its description), \emph{add} a new cluster, \emph{remove} a cluster, or \emph{no\_change} when no modification is warranted. Each recommendation carries the supporting evidence used to derive it --- the search queries issued, the text identifiers inspected, and the investigator's reasoning --- together with a confidence level. A \emph{no\_change} recommendation additionally records the hypothesis that was rejected, so that subsequent agents do not re-investigate the same question.

The investigator's recommendation is passed back to the orchestrator, which decides whether to implement it. Structural changes (\emph{merge}, \emph{split}, \emph{add}, \emph{remove}) invalidate the existing audit metrics, so an audit pass is naturally re-triggered to validate the modified taxonomy on fresh evidence.

\subsection{Critic}
\label{sec:method-critic}
The critic is dispatched once the taxonomy has been audited, and again after any major change to it. Whereas the auditor asks whether the existing clusters can account for previously unseen texts, the critic evaluates the taxonomy against a fixed six-axis checklist, scoring each axis as a pass, warning, or failure:
\begin{itemize}
  \setlength{\itemsep}{0pt}\setlength{\parsep}{0pt}
\item \emph{Overlap}: are any two clusters redundant?
\item \emph{Gaps}: are any categories left uncovered?
\item \emph{Granularity}: are some clusters substantially broader or narrower than others?
\item \emph{Boundaries}: can similar clusters be told apart, including on boundary-case texts?
\item \emph{Descriptions}: do they define general concepts, not overfit to specific evidence?
\item $k$-\emph{range}: is the count inside $[k_{\min}, k_{\max}]$?
\end{itemize}
For axes that require empirical evidence, the critic pulls targeted texts from the corpus: similarity-based retrieval to probe boundary cases between similar-looking clusters, and audit-based retrieval to inspect unclustered patterns when looking for gaps. The user task description, where supplied, defines what counts as a good cluster set: a critique that the clusters are technically coherent but ill-suited to the user's stated purpose is itself a critical finding. The critic's output is a ranked list of issues.

The critic's findings drive the orchestrator's subsequent dispatches. \emph{Overlap}, \emph{granularity}, \emph{boundary} and \emph{gap} issues typically become questions for the investigator; \emph{description}-quality issues prompt the orchestrator to rewrite the affected descriptions directly, without dispatching another agent; and $k$-\emph{range} violations may prompt a fresh synthesis pass. The orchestrator finalises the taxonomy once the critic raises no blocking issues, audit coverage exceeds 85\%, and every cluster has high confidence (a mean of at least 4 out of 5), or once the last two or three actions have not improved it; there is no fixed limit on the number of dispatches.

\begin{table*}[t]
  \centering
  \small
  \begin{tabular}{lllrrrr}
    \toprule
    \textbf{Dataset} & \textbf{Data type} & \textbf{Cluster type} & $N$ & $k$ & \textbf{Not in cluster} & \textbf{Mean chars} \\
    \midrule
    Banking77 & banking queries & intent & 3,080 & 77 & 0.0\% & 54 \\
    CLINC150 & assistant queries & intent & 5,500 & 150 & 18.2\% & 40 \\
    MASSIVE-Intent & assistant queries & intent & 2,974 & 60 & 0.0\% & 35 \\
    MASSIVE-Domain & assistant queries & domain & 2,974 & 18 & 0.0\% & 35 \\
    GoEmotions & Reddit comments & emotion & 45,446 & 27 & 35.3\% & 67 \\
    20 Newsgroups & Usenet posts & topic & 18,331 & 20 & 0.0\% & 1200 \\
    StackExchange & forum titles & topic (forum) & 4,156 & 121 & 0.0\% & 57 \\
    \bottomrule
  \end{tabular}
  \caption{Descriptive statistics of benchmark datasets.}
  \label{tab:datasets}
\end{table*}

\subsection{Classification}
\label{sec:method-classification}

Once the taxonomy is final, the cluster definitions are applied to the full corpus. Each text in $D$ is sent independently to a language model together with a structured-output schema that constrains the response to one of the taxonomy's cluster identifiers or the special value \emph{none}; the model returns this identifier together with a confidence score and a short reasoning trace.

The classification system prompt is built once per run from the finalised taxonomy. It contains each cluster's name and description but not the example texts that were collected as evidence during discovery, so the classifier must generalise from the definitions alone rather than retrieving observed instances. The system prompt is held constant across every text and cached, so the per-text inference cost is dominated by the short user message containing the text under classification; classification can be run synchronously when latency matters or via a batch interface when cost matters more.

An optional refinement step allows the classification prompt to be tuned against a small human-labelled subset of the corpus. A handful of variant prompt headers are scored by classifying the labelled subset under each header and computing agreement with the human labels; the best-scoring header is retained as the classification prompt for the full-corpus pass.

\section{Datasets}
\label{sec:datasets}

We evaluate on seven public text-clustering benchmarks (Table~\ref{tab:datasets}): Banking77 \citep{casanueva-etal-2020-banking77}, CLINC150 \citep{larson-etal-2019-clinc150}, MASSIVE-Intent and MASSIVE-Domain \citep{fitzgerald-etal-2023-massive}, GoEmotions \citep{demszky-etal-2020-goemotions}, 20 Newsgroups \citep{lang1995newsweeder}, and StackExchange \citep{zhang-etal-2023-clusterllm}. Six are short-text corpora --- banking and assistant queries, Reddit comments, and forum titles --- that recent LLM-based methods have re-purposed for clustering \citep{zhang-etal-2023-clusterllm}; 20 Newsgroups adds long documents. Two of the seven include a gold ``none'' class --- CLINC's out-of-scope queries and GoEmotions' neutral category --- which we deliberately retain so that the benchmark can credit methods that natively express ``no cluster.'' Appendix~\ref{sec:appendix-datasets} gives each dataset's exact source, config, split, and cleaning recipe.

\section{Experimental setup}
\label{sec:experiments}

\paragraph{Baseline selection.} We compare against seven prior methods. We choose LDA \citep{blei2003lda} as a representative member of probabilistic topic models. For dense embedding models, we choose Sentence-BERT \citep{reimers-gurevych-2019-sentence-bert} + $k$-means, and BERTopic \citep{grootendorst2022bertopic}. We also use LLM embeddings + $k$-means \citep{petukhova2024llm}, and ClusterLLM \citep{zhang-etal-2023-clusterllm}, as LLM-based dense embedding refinements. For LLM-driven taxonomy generation methods, we compare with TopicGPT \citep{pham-etal-2024-topicgpt} and \citet{huang-he-2024-text-clustering}. We omit TnT-LLM \citep{wan2024tntllm}, whose code is unreleased. 

For all cases using LLMs, for a fair comparison, we use a \emph{frontier tier} (Claude Opus 4.7) for heavy-weight tasks that run on small samples of the dataset, such as taxonomy generation, merging and refinement, and a \emph{cheap tier} (GPT-5-mini) for light-weight tasks over the whole corpus, such as classification into the taxonomy. Therefore we reimplement all methods directly. Per-method implementation settings are presented in Appendix~\ref{sec:appendix-impl}.

\paragraph{$k$ handling.} Some methods require the number of clusters, $k$, to be passed as an input, while others discover $k$. For methods that require $k$, we pass the true $k$. We report results separately for given-$k$ and discover-$k$ methods. Our method and BERTopic can handle both, so we report both. For discover-$k$, our headline discovery setting, we set a $\pm20\%$ window around the gold count, rounded to integers, which supplies approximate granularity while leaving room to over- or under-estimate within the range. Because this range is gold-derived, we additionally report a fully unsupervised \emph{no-$k$} setting that provides no gold-derived interval at all --- the method sees only the corpus and the optional free-text lens and discovers $k$ from the data --- putting it on the same footing as TopicGPT and Huang \& He, which receive no such range.

\paragraph{Noise / out-of-cluster handling.} Some datasets include a noise/out-of-cluster class, which is removed by some papers in their evaluation. We include it, as this better represents real-world use. For the methods that can handle it natively, we use the no-none cluster mode when there is no gold none class, and the with-none cluster mode when there is a gold none class.

\paragraph{Document-length cap.} For every LLM-using method each call that passes a document body to a language model truncates the text to 512 tokens. The cap binds only on 20~Newsgroups (9.4\% documents) and a single GoEmotions comment.

\paragraph{Seeds and sampling.} Baselines are run with a single seed, owing to the cost of the full sweep. Our method is additionally run at three seeds in the discover-$k$ setting, resampling every stochastic stage; Appendix~\ref{sec:appendix-seeds} reports mean and standard deviation per dataset and confirms the ranking is unchanged under every draw.

\paragraph{Gold set leakage.} For two methods, TopicGPT and Huang \& He, the headline reported results are computed by using a sample of the gold set in the prompts. This helps inform the granularity of the clusters produced. However, this is tantamount to gold set leakage. Therefore we run these methods without prompting with the gold set (referred to as the `0\% seed' configuration in the original paper).

\paragraph{Metrics.} We report Adjusted Rand Index (ARI), Normalised Mutual Information (NMI), and clustering accuracy under Hungarian assignment (ACC).\footnote{ARI and NMI use the standard scikit-learn implementations; ACC is computed via \texttt{scipy.optimize.linear\_sum\_assignment} on the predicted$\times$gold contingency matrix.} The three metrics disagree in informative ways on datasets with a gold ``none'' class: NMI tolerates a method redistributing ``none'' documents across in-scope clusters; ARI and ACC penalise it.

\ifdefined\acdeckone\else\newsavebox{\acdeckone}\fi
\ifdefined\acdecktwo\else\newsavebox{\acdecktwo}\fi
\begin{table*}[t]
  \centering
  \sbox{\acdeckone}{%
  \begin{tabular}{l cccc cccc cccc cccc cccc}
    \toprule
     & \multicolumn{4}{c}{\textbf{Banking77}} & \multicolumn{4}{c}{\textbf{CLINC150}} & \multicolumn{4}{c}{\textbf{MASSIVE-Intent}} & \multicolumn{4}{c}{\textbf{MASSIVE-Domain}} & \multicolumn{4}{c}{\textbf{GoEmotions}} \\
    \cmidrule(lr){2-5}\cmidrule(lr){6-9}\cmidrule(lr){10-13}\cmidrule(lr){14-17}\cmidrule(lr){18-21}
    \textbf{Method} & $\hat{k}$ & ARI & NMI & ACC & $\hat{k}$ & ARI & NMI & ACC & $\hat{k}$ & ARI & NMI & ACC & $\hat{k}$ & ARI & NMI & ACC & $\hat{k}$ & ARI & NMI & ACC \\
    \midrule
    \multicolumn{21}{l}{\textit{Given $k$}} \\
    \midrule
    LDA & 77 & 0.20 & 0.55 & 0.30 & 150 & 0.06 & 0.56 & 0.30 & 60 & 0.17 & 0.44 & 0.29 & 18 & 0.17 & 0.26 & 0.31 & 27 & 0.01 & 0.04 & 0.10 \\
    SBERT+$k$-means & 77 & 0.55 & 0.82 & 0.65 & 150 & 0.21 & 0.83 & 0.66 & 60 & 0.39 & 0.71 & 0.52 & 18 & 0.42 & 0.63 & 0.56 & 27 & 0.02 & 0.07 & 0.11 \\
    BERTopic & 77 & 0.56 & 0.83 & 0.64 & 150 & 0.26 & 0.82 & 0.65 & 60 & 0.45 & 0.73 & 0.58 & 18 & 0.48 & 0.64 & 0.61 & 27 & 0.01 & 0.03 & 0.24 \\
    LLM-embedding+$k$-means & 77 & 0.62 & 0.85 & 0.72 & 150 & 0.28 & 0.87 & 0.70 & 60 & 0.40 & 0.73 & 0.51 & 18 & 0.49 & 0.70 & 0.62 & 27 & 0.04 & 0.13 & 0.17 \\
    ClusterLLM & 77 & 0.60 & 0.84 & 0.70 & 150 & 0.21 & 0.85 & 0.68 & 60 & 0.47 & 0.77 & 0.59 & 18 & 0.44 & 0.68 & 0.59 & 27 & 0.06 & 0.21 & 0.25 \\
    Agentic clustering (ours) & 77 & \underline{0.73} & \underline{0.89} & \underline{0.84} & 150 & \underline{0.73} & \underline{0.93} & \underline{0.89} & 60 & \underline{0.74} & \underline{0.86} & \underline{0.85} & 18 & \underline{0.73} & \underline{0.80} & \underline{0.82} & 27 & \underline{0.08} & \underline{0.24} & \underline{0.35} \\
    \midrule
    \multicolumn{21}{l}{\textit{Discover $k$}} \\
    \midrule
    BERTopic & 90 & 0.60 & 0.84 & 0.69 & 166 & 0.26 & 0.82 & 0.70 & 67 & 0.43 & 0.73 & 0.57 & 67 & 0.27 & 0.64 & 0.36 & 429 & 0.01 & 0.09 & 0.22 \\
    TopicGPT & 5 & 0.01 & 0.16 & 0.04 & 14 & 0.05 & 0.46 & 0.15 & 15 & 0.10 & 0.47 & 0.31 & 12 & 0.18 & 0.45 & 0.43 & 27 & 0.00 & 0.04 & 0.11 \\
    Huang \& He & 131 & 0.57 & 0.84 & 0.64 & 766 & 0.13 & 0.83 & 0.59 & 172 & 0.37 & 0.77 & 0.48 & 476 & 0.12 & 0.61 & 0.17 & 207 & 0.03 & 0.17 & 0.16 \\
    Agentic clustering, $k\pm20\%$ (ours) & 83 & 0.65 & 0.87 & \underline{0.76} & 151 & \underline{0.81} & \underline{0.94} & \underline{0.93} & 64 & 0.55 & 0.81 & 0.65 & 20 & \underline{0.62} & \underline{0.76} & \underline{0.71} & 25 & \underline{0.08} & \underline{0.24} & 0.31 \\
    Agentic clustering, no $k$ (ours) & 72 & \underline{0.66} & \underline{0.88} & 0.75 & 177 & 0.33 & 0.90 & 0.80 & 57 & \underline{0.70} & \underline{0.85} & \underline{0.83} & 16 & 0.59 & 0.73 & 0.71 & 22 & 0.08 & 0.23 & \underline{0.31} \\
    \bottomrule
  \end{tabular}%
  }
  \sbox{\acdecktwo}{%
  \begin{tabular}{l cccc cccc rrrr rr}
    \toprule
     & \multicolumn{4}{c}{\textbf{20 Newsgroups}} & \multicolumn{4}{c}{\textbf{StackExchange}} & \multicolumn{4}{c}{\textbf{Tokens (M)}} & \multicolumn{2}{c}{\textbf{Spend (\$)}} \\
    \cmidrule(lr){2-5}\cmidrule(lr){6-9}\cmidrule(lr){10-13}\cmidrule(lr){14-15}
    \textbf{Method} & $\hat{k}$ & ARI & NMI & ACC & $\hat{k}$ & ARI & NMI & ACC & Frontier in & Frontier out & Cheap in & Cheap out & Frontier & Cheap \\
    \midrule
    \multicolumn{15}{l}{\textit{Given $k$}} \\
    \midrule
    LDA & 20 & 0.19 & 0.36 & 0.34 & 121 & 0.01 & 0.31 & 0.08 & -- & -- & -- & -- & -- & -- \\
    SBERT+$k$-means & 20 & 0.42 & 0.58 & 0.61 & 121 & 0.16 & 0.52 & 0.29 & -- & -- & -- & -- & -- & -- \\
    BERTopic & 20 & 0.31 & 0.53 & 0.39 & 70 & 0.17 & 0.51 & 0.30 & -- & -- & -- & -- & -- & -- \\
    LLM-embedding+$k$-means & 20 & \underline{0.49} & 0.63 & 0.64 & 121 & 0.34 & 0.67 & 0.47 & -- & -- & -- & -- & -- & \$0.39 \\
    ClusterLLM & 20 & 0.48 & 0.64 & 0.67 & 121 & 0.31 & 0.66 & 0.45 & -- & -- & -- & -- & -- & \$1.53 \\
    Agentic clustering (ours) & 20 & 0.46 & \underline{0.68} & \underline{0.72} & 121 & \underline{0.43} & \underline{0.73} & \underline{0.59} & 124 & 3.8 & 216 & 16 & \$100 (\$224) & \$22 \\
    \midrule
    \multicolumn{15}{l}{\textit{Discover $k$}} \\
    \midrule
    BERTopic & 272 & 0.17 & 0.51 & 0.20 & 70 & 0.17 & 0.51 & 0.30 & -- & -- & -- & -- & -- & -- \\
    TopicGPT & 18 & 0.28 & 0.53 & 0.39 & 30 & 0.13 & 0.54 & 0.28 & 5.4 & 0.17 & 100 & 43 & \$31 & \$66 \\
    Huang \& He & 377 & 0.22 & 0.53 & 0.29 & 2764 & 0.02 & 0.69 & 0.10 & 0.88 & 0.03 & 384 & 33 & \$5 & \$44 \\
    Agentic clustering, $k\pm20\%$ (ours) & 22 & \underline{0.58} & \underline{0.67} & \underline{0.73} & 117 & \underline{0.46} & \underline{0.75} & \underline{0.61} & 173 & 4.9 & 265 & 17 & \$100 (\$292) & \$22 \\
    Agentic clustering, no $k$ (ours) & 18 & 0.53 & 0.67 & 0.71 & 35 & 0.18 & 0.61 & 0.30 & 175 & 4.2 & 260 & 18 & \$100 (\$265) & \$24 \\
    \bottomrule
  \end{tabular}%
  }
  \resizebox{\fpeval{\wd\acdeckone/max(\wd\acdeckone,\wd\acdecktwo)}\textwidth}{!}{\usebox{\acdeckone}}

  \vspace{0.5em}

  \resizebox{\fpeval{\wd\acdecktwo/max(\wd\acdeckone,\wd\acdecktwo)}\textwidth}{!}{\usebox{\acdecktwo}}
  \caption{Clustering results across seven benchmarks. Underlined values are the best in each column within the given-$k$ and discover-$k$ blocks. $\hat{k}$ is the number of clusters produced. The token columns report billed frontier-tier (Claude Opus 4.7) and cheap-tier (GPT-5-mini) input and output in millions, summed across the seven datasets; for our method, frontier input includes the context its agents re-send on every turn. TopicGPT's and Huang \& He's frontier tokens are estimated from their saved prompts and responses, adjusted for Opus 4.7's tokenizer. Spend is the total USD across the seven datasets, split by tier, at metered API rates; our method's frontier spend is instead shown as the flat \$100 Claude Code Max subscription it is designed to run under, with the metered equivalent in brackets.}
  \label{tab:results}
\end{table*}

\section{Results}
\label{sec:results}

Table~\ref{tab:results} reports the results of the experiments. Our method is state-of-the-art on six of seven datasets; on 20~Newsgroups it trails LLM-embedding+$k$-means on ARI (0.46 vs.\ 0.49) while leading on NMI and ACC. On five of the seven the ARI margin over the second-best method exceeds 25\%. Appendix~\ref{sec:appendix-qualitative} complements these metrics with a qualitative analysis of the method's output, including where and why it disagrees with the gold labels.

\paragraph{Other LLM taxonomy models struggle to control granularity.} Without the gold-set injection, both Huang \& He and TopicGPT struggle to control the granularity of the clusters produced. Huang \& He produces a large number of small clusters, while TopicGPT produces a small number of large clusters, leading them to perform worse than LDA on some datasets. In both cases, their fixed prompts ask the LLM to ``categorize'' inputs but contain no signal about the abstraction level at which categories should be drawn --- and because the surrounding pipeline is programmatic, there is no inner loop that could adapt that signal from the data. The MASSIVE-Intent / MASSIVE-Domain pair is the natural experiment that exposes this: the two datasets share the same 2,974 utterances and differ only in granularity and clustering intent, yet both TopicGPT and Huang \& He are provided with the exact same prompt in both cases. This highlights the importance of user-given clustering intents, which is explored further in Section~\ref{sec:ablations}.

\paragraph{No-$k$ discovery.} With no gold-derived range at all, the method beats every discover-$k$ baseline on ARI on all seven datasets, and matches or beats them on ACC; the only shortfall is StackExchange NMI, where Huang \& He scores higher. The range is therefore not the source of the method's advantage over prior work. It is nonetheless a useful granularity control, worth $+0.10$ ARI averaged over the seven datasets.

\paragraph{Dense embedding models.} LLM-embedding+$k$-means beats LDA, SBERT+$k$-means, and BERTopic on every dataset on the majority of metrics, in line with the findings of \citet{petukhova2024llm}. ClusterLLM's triplet fine-tuning closes most of that gap, sometimes more. 

\paragraph{Ablations.}\label{sec:ablations} We ablate two components of our method in the discover-$k$ setting (Table~\ref{tab:ablations}, Appendix~\ref{sec:appendix-ablations}). First, we run our method without the optional user cluster intent. Removing it costs $0.05$ ARI on average across the seven datasets, though the effect is uneven. The intent helps most where a corpus admits several valid taxonomies, rather than uniformly.

Second, we remove the post-synthesis refinement loop (auditor, critic, investigator). In the tuned configuration this costs only $0.01$ ARI, because the multi-view proposer and synthesizer already produce a strong first pass that leaves the loop little to correct. That is not evidence the loop is idle: its contribution scales with how imperfect the first pass is. The design ablations in Appendix~\ref{sec:appendix-ablations} show this directly. When each proposer's sample is halved, the loop corrects seven times as much cluster-count error as it does at the default sample size (38 versus 5 points of $\Sigma|\Delta k|$), and dispatches a third more investigation rounds to do it (7.9 versus 5.9). The degraded front-end then ends up no worse than the healthy one ($\Sigma|\Delta k|$ 19 versus 21, Table~\ref{tab:design-ablations}). A fixed-schedule pipeline cannot make that trade. The front-end and the loop are complementary: the front-end does most of the work when the corpus is easy, and the loop recovers quality when it is not. The same ablations show the front-end is itself well-behaved: the number of proposers traces a clean optimum at six to seven.

\paragraph{Cost and efficiency.} The spend columns of Table~\ref{tab:results} report each method's total spend across the seven datasets, split by model tier. Three methods --- ours, TopicGPT, and Huang \& He --- use a frontier-tier model (Claude Opus 4.7) to build their taxonomies and a cheap tier (GPT-5-mini) for full-corpus classification. The classical baselines (LDA, SBERT+$k$-means, BERTopic) run locally at no API cost, and the metered embedding baselines are inexpensive (LLM-embedding+$k$-means \$0.39, ClusterLLM \$1.53). Comparing costs across the three LLM-taxonomy methods is not straightforward. Our method is designed to run as a Claude Code plugin, where its agent loop is covered by the flat Claude Code Max subscription ($\sim$\$100 per month), whereas the baselines issue discrete prompts that can equally be billed per token. We therefore report our frontier spend both as the subscription and, in brackets, at metered API rates.

Given a Claude Code subscription, our method's better results come at no added cost over the other LLM-taxonomy methods. Beyond the flat fee, its only spend is the cheap-tier classification pass, and this is the lowest of the three: \$22, against \$44 for Huang \& He and \$66 for TopicGPT. The gap is likely to widen on corpora larger than these benchmarks (2,974 to 45,446 documents), since our frontier-tier usage does not grow with the corpus and our cheap-tier spend is the lowest. Without a subscription, however, the frontier tier would cost \$224--\$292 across the seven datasets at metered API rates, against about \$31 for TopicGPT and \$5 for Huang \& He. The method should therefore only be run inside Claude Code.

Table~\ref{tab:results} also reports the billed tokens for each tier. Our method's frontier usage does not grow with the corpus: the agent loop reads only fixed-size samples, so per dataset it stays between 9.4M and 24.8M tokens across corpora spanning 2,974 to 45,446 documents ($r=-0.21$ with corpus size), scaling instead with the number of clusters ($r=+0.80$). Only the cheap, batchable classification pass scales with dataset size. Huang \& He's frontier call, by contrast, grows with the corpus, since it merges the labels generated across every document.

\section{Conclusion}
We have presented an agentic approach to text clustering, in which an orchestrator dispatches a small set of specialised LLM agents --- proposer, synthesizer, auditor, investigator, critic --- in a state-dependent order rather than executing the fixed pipeline of prior LLM-driven taxonomy work. Our method is state-of-the-art on six of seven datasets, and on five of them is more than 25\% better than the second-best method on ARI.

\section*{Limitations}

\paragraph{Re-implementation of baselines.} Every method was re-implemented while holding the use of privileged information (such as gold-label seeding) and LLMs consistent across methods. We have tried to keep these re-implementations as faithful as possible --- preserving each method's prompts, hyperparameters, and pipeline structure --- and have documented every deviation in Appendix~\ref{sec:appendix-impl}. Even so, perfect comparability across methods built around very different LLM workflows is not achievable, and small cross-method differences in score may reflect implementation choices rather than the underlying methods.

\paragraph{Baseline seeds.} Owing to the cost of the full sweep --- seven datasets across eight methods, several of which make per-document LLM calls --- every baseline number is a single-seed run. We characterise our own method's variance instead (Appendix~\ref{sec:appendix-seeds}): every one of our three runs beats the strongest discover-$k$ baseline on every dataset and on all three metrics, so the comparison does not rest on a fortunate draw of ours. Several baselines are themselves stochastic, however, and we have not measured their spread; a favourable baseline draw could narrow our two closest margins, on Banking77 and MASSIVE-Intent.

\paragraph{English only.} All seven benchmarks are English. We do not evaluate on non-English or code-mixed corpora and make no claim that the agentic pipeline's quality lead transfers to other languages.

\paragraph{Closed-model dependence.} The method depends on closed models (Claude Opus 4.7 via Claude Code, GPT-5-mini via the OpenAI Batch API), which limits reproducibility, locks users to specific vendors, and means behaviour can shift silently as those models are updated. Its cost also depends on a Claude Code subscription: on metered API billing its frontier tier is several times more expensive than the baselines' (\S\ref{sec:results}). Open-weight substitution is possible in principle --- the orchestrator and per-agent prompts are model-agnostic --- but is unevaluated here.

\paragraph{Risks.}
Our method inherits three risks worth flagging. First, taxonomies discovered by the pipeline may be applied to consequential downstream decisions --- routing support tickets, triaging moderation queues, categorising sensitive documents --- and misclassifications in those settings carry material costs that are not captured by the aggregate metrics we report. Second, every agent in the loop is an LLM call, so whatever biases the underlying models carry --- in what they treat as a natural category, what they label as noise, how they handle minority dialects or non-Western contexts --- propagate into the resulting taxonomies, which is particularly relevant for corpora about people. Third, the user-supplied task description is a feature for legitimate analysis but also lowers the barrier to partitioning a corpus along any axis a user names (e.g.\ political affiliation or demographic markers), amplifying a dual-use concern that is not novel to this work but is made more accessible by it.

\section*{Acknowledgments}

The method described in this paper is itself a multi-agent system built on top of frontier large language models (Claude Opus 4.7 and GPT-5-mini); the use of generative AI in the method is the central methodological contribution and is documented throughout the paper. In addition, we disclose two further uses of generative AI outside the method itself. First, Claude was used as a writing assistant to draft, restructure, and copy-edit portions of this manuscript; all scientific claims, experimental decisions, and the final text are the authors' responsibility. Second, Claude Code was used to assist in re-implementing the prior LLM-based baselines (TopicGPT, Huang \& He, ClusterLLM) from their authors' released code, under the comparability constraints documented in Appendix~\ref{sec:appendix-impl}; every re-implementation was reviewed by the authors against the upstream source.

\bibliography{references}

\appendix
\setcounter{dbltopnumber}{2} %
\raggedbottom

\twocolumn[{%
\setlength{\columnwidth}{\textwidth}\setlength{\linewidth}{\textwidth}\setlength{\hsize}{\textwidth}%
\begin{algorithm}[H]
\caption{Agentic clustering.}
\label{alg:orchestrator}
\small
\begin{algorithmic}[1]
\Require corpus $D$, $k$-range $[k_{\min}, k_{\max}]$, task description $T$ (optional)
\Ensure taxonomy $C$, and a cluster or \emph{none} for every text in $D$
\State $S \gets$ empty workspace (drafts, taxonomy, audits, critiques, seen set)
\Loop
\Statex \hspace{\algorithmicindent}\textit{(no limit on the number of iterations)}
  \State $a \gets \textsc{Orchestrator}(S, T, [k_{\min}, k_{\max}])$, the next action, chosen from the current workspace
  \If{$a = \textsc{Propose}$}
    \State dispatch proposers in parallel, each with a style directive, on samples of unseen texts; add their draft cluster sets to $S$
    \State add the sampled texts to the seen set
  \ElsIf{$a = \textsc{Synthesize}$}
    \State $C \gets$ \textsc{Synthesizer}(draft cluster sets in $S$, $T$, $[k_{\min}, k_{\max}]$)
  \ElsIf{$a = \textsc{Audit}$}
    \State \textsc{Auditor} assigns each text in a random or stratified sample of unseen texts to a cluster in $C$ or \emph{none}
    \State add the sampled texts to the seen set; record coverage and per-cluster confidence in $S$
    \State flag clusters whose mean confidence is below 3 as weak
  \ElsIf{$a = \textsc{Investigate}(q)$, where $q$ is a question about $C$ (e.g.\ should two clusters merge?)}
    \State $r \gets \textsc{Investigator}(q, D, C, T)$, a recommendation to merge, split, rename, add, remove, or make no change
    \State apply $r$ to $C$ if the orchestrator accepts it
  \ElsIf{$a = \textsc{Critique}$}
    \State \textsc{Critic} evaluates $C$ against $T$ and $[k_{\min}, k_{\max}]$ for overlap, gaps, granularity, boundaries, descriptions, and $k$-range
    \State add the critic's issues to $S$
  \ElsIf{$a = \textsc{Rewrite}$}
    \State the orchestrator rewrites the descriptions the critic flagged
  \ElsIf{$a = \textsc{Finalize}$}
\Statex \hspace{\algorithmicindent}\hspace{\algorithmicindent}\textit{(chosen once the critic raises no blocking issues, audit coverage exceeds 85\%, and every cluster's}
\Statex \hspace{\algorithmicindent}\hspace{\algorithmicindent}\textit{mean audit confidence is at least 4, or once the last two or three actions have not improved $C$)}
    \If{every cluster in $C$ has $\geq 5$ audit assignments}
      \State \textbf{break}
    \EndIf
  \EndIf
\EndLoop
\State \Return $C$ and $\textsc{Classify}(D, C)$
\end{algorithmic}
\end{algorithm}
\vspace{1em}}]
\section{Pseudocode}
\label{sec:appendix-orchestrator}

\noindent Algorithm~\ref{alg:orchestrator} provides pseudocode for the agentic clustering method. It is not a fixed pipeline: at each step the orchestrator chooses the next action from the current state of the workspace.

\section{Dataset processing details}
\label{sec:appendix-datasets}

All loaders are deterministic given a fixed scikit-learn version (for 20 Newsgroups) and the Hugging Face dataset revisions recorded in each dataset's \texttt{meta.json}.

Each dataset is used for the intent classification, emotion classification, or topic clustering purpose for which it was originally released, which aligns with our text-clustering evaluation; per-dataset license terms are documented at the end of each subsection below.

\subsection{Banking77}

Banking77 \citep{casanueva-etal-2020-banking77} consists of customer-service queries to an online bank, labelled with one of 77 fine-grained intents (e.g.\ \emph{activate\_my\_card}, \emph{pending\_top\_up}). We use the 3,080-example test split unmodified, matching prior LLM-clustering work. The dataset is loaded from \texttt{PolyAI/banking77} on Hugging Face Hub via the \texttt{refs/} \texttt{convert/parquet} revision. The \texttt{ClassLabel} feature provides the canonical label names, which we copy directly into the taxonomy. \emph{License:} CC-BY-4.0, consistent with research re-use.

\subsection{CLINC150}
\label{sec:datasets-clinc}

CLINC150 \citep{larson-etal-2019-clinc150} is a 150-intent benchmark with a dedicated out-of-scope (OOS) class for queries outside the domain. We use the \texttt{plus} config's 5,500-example test split (4,500 in-scope + 1,000 OOS). Unlike some prior LLM-clustering work, which discards OOS examples, we retain them and evaluate them as the gold ``none'' class; methods without native ``none'' support are expected to score worse here. The dataset is loaded from \texttt{clinc/clinc\_oos} on Hugging Face Hub. The dataset's parquet conversion does not preserve config names as Hugging Face dataset configs; instead, each config lives as a subdirectory of the \texttt{refs/convert/parquet} branch. We load directly from \nolinkurl{hf://datasets/clinc/clinc_oos@refs/convert/parquet/plus/test/0000.parquet}. Examples with source intent ``oos'' are remapped to \texttt{gold\_label\_id} $=-1$ and \texttt{gold\_label\_name} $=$ \texttt{\_\_none\_\_}; the remaining 150 in-scope intent ids are compacted to a contiguous $0\ldots 149$ range. Two prior LLM-clustering papers use different CLINC configurations and both drop the OOS examples: \citet{zhang-etal-2023-clusterllm} use the same \texttt{plus} test split as we do but discard OOS, while \citet{viswanathan2024fewshot} use the \texttt{small} configuration. We use \texttt{plus} so that the OOS subset is available, and retain it as the gold ``none'' class. \emph{License:} CC-BY-3.0, consistent with research re-use.

\subsection{MASSIVE-Intent and MASSIVE-Domain}

MASSIVE-Intent and MASSIVE-Domain \citep{fitzgerald-etal-2023-massive} share the same 2,974 EN test utterances under two different label projections: 60 fine-grained intents and 18 coarse scenarios respectively. This pair isolates the effect of taxonomy granularity from corpus content. The two splits are loaded from \texttt{mteb/amazon\_massive\_intent} and \texttt{mteb/amazon\_massive\_scenario} respectively, both using the \texttt{en} configuration and the \texttt{test} split. In both datasets the gold label ships as a human-readable string (e.g.\ \emph{alarm\_set}, \emph{calendar}); we assign integer ids in sorted-string order to obtain a stable mapping. To keep the gold label space exactly $k{=}60$ for MASSIVE-Intent (per the MASSIVE paper) even though the test split is missing one of the 60 intents (\emph{cooking\_query}), the canonical label set is derived from the union of train, validation, and test labels --- only test rows are then emitted as documents. MASSIVE-Domain uses an analogous union to fix $k{=}18$. The processing pipeline cross-checks that the EN test texts in MASSIVE-Intent and MASSIVE-Domain coincide row-by-row, since the two MTEB datasets are different label projections of the same corpus and any drift would silently break the comparison. \emph{License:} CC-BY-4.0, consistent with research re-use.

\subsection{GoEmotions}
\label{sec:datasets-goemotions}

GoEmotions \citep{demszky-etal-2020-goemotions} comprises Reddit comments labelled with 27 emotions plus a neutral class. Like with CLINC's OOS, we retain neutral as the gold ``none'' class, departing from prior work that drops it \citep{zhang-etal-2023-clusterllm}. The dataset is loaded from \texttt{google-research-datasets/go\_emotions} on Hugging Face Hub, using the \texttt{simplified} configuration. The \texttt{simplified} config restricts to single-label rows but ships them as a list (\texttt{labels: list[int]} of length one); rows whose list length is not one are skipped (8,817 multi-label rows). The remaining 45,446 single-label documents are concatenated across train, validation, and test --- the same recipe used by \citet{zhang-etal-2023-clusterllm}. \emph{Neutral} (source label id 27) is remapped to \texttt{gold\_label\_id} $=-1$ and \texttt{gold\_label\_name} $=$ \texttt{\_\_none\_\_}, and the remaining 27 in-scope emotion ids are compacted to $0\ldots 26$. \emph{License:} Apache-2.0, consistent with research re-use.

\subsection{20 Newsgroups}

20 Newsgroups \citep{lang1995newsweeder} is a long-form topic-clustering benchmark of Usenet posts across 20 topics. We adopt the BERTopic preprocessing recipe \citep{grootendorst2022bertopic} via scikit-learn (version 1.8.0) using \texttt{fetch\_20newsgroups(subset='all', remove=('headers',} \texttt{'footers',} \texttt{'quotes'))}. The \texttt{remove} flags strip the header block, the signature/footer block, and quoted reply text --- all three contain newsgroup names or quote markers that would otherwise leak the gold label. After stripping we drop rows that become empty whitespace (515 documents), leaving 18,331 documents across the 20 newsgroup labels. We use scikit-learn rather than \texttt{SetFit/20\_newsgroups} on Hugging Face Hub because the latter does not expose the \texttt{remove} parameter. We do not use MTEB's \texttt{twentynewsgroups-clustering}, which is subject-line only ($\sim$32 chars on average) and would not exercise the long-document regime that motivates this dataset's inclusion. \emph{License:} the 20 Newsgroups corpus is in the public domain and has been freely redistributed for research for three decades, including by scikit-learn under its BSD-3-Clause distribution; our use is consistent with this.

\subsection{StackExchange}
\label{sec:appendix-stackexchange}

StackExchange consists of 4,156 forum question titles drawn from 121 stackexchange.com sites, where the site identity (e.g.\ \emph{english}, \emph{history}) serves as the gold cluster label. We use the \texttt{small} subset released by \citet{zhang-etal-2023-clusterllm}, loaded from the data zip linked in the \texttt{ClusterLLM} GitHub repository's README. The loader auto-downloads the zip to \texttt{data/raw/clusterllm/} on first run and reads \texttt{datasets/stackexchange/small.jsonl}. We strip the \texttt{.txt} suffix that ClusterLLM's recipe appends to the site names, yielding 121 in-scope labels. We adopt this resource rather than Hugging Face's \texttt{mteb/stackexchange-clustering} dataset because the latter is structured as a list of 25 clustering subsets (5,010--26,666 documents each, 121 distinct labels in total), and MTEB's reported score for this task is the mean V-measure across per-subset $k$-means runs. That protocol is incompatible with our single-partition ARI/NMI/ACC reporting on a fixed taxonomy. The ClusterLLM-small file is the de facto small-StackExchange benchmark in the LLM-clustering line of work, and at $n{=}4{,}156$ sits inside the size range of our other six datasets. \emph{License:} the underlying Stack Exchange user content is CC-BY-SA 4.0 (Stack Exchange's site-wide content licence); the ClusterLLM repackaging is released alongside the paper without an explicit licence for the bundle, and we use it under the same research-reproduction terms applied by prior work \citep{zhang-etal-2023-clusterllm}.

\section{Implementation details}
\label{sec:appendix-impl}

\paragraph{Compute and infrastructure.} The dominant compute cost in our pipeline and in the four LLM-using baselines (LLM-embedding+$k$-means, ClusterLLM, TopicGPT, and Huang \& He) is consumed by closed-model API calls (Claude Opus 4.7, GPT-5-mini, and OpenAI \texttt{text-embedding-3-large}); we account for this as USD spend in the Cost paragraph of \S\ref{sec:results} and the final column of Table~\ref{tab:results} rather than in GPU hours. Local compute is light: LDA, the SBERT-based pipelines (SBERT+$k$-means, BERTopic, and ClusterLLM's Instructor encoder at inference), and the per-dataset processing all run on a single workstation in minutes per dataset. The only non-workstation step is ClusterLLM's perspective fine-tuning (the upstream \texttt{finetune.sh}), which runs on a single NVIDIA A100 for approximately ten minutes per dataset. The open-weight encoders used in this study are \texttt{sentence-transformers/all-mpnet-base-v2} (110M parameters) and \texttt{hkunlp/instructor-large} (335M parameters); the closed models we call (Claude Opus 4.7, GPT-5-mini, \texttt{text-embedding-3-large}) do not publish parameter counts.

\paragraph{LDA.} scikit-learn's \texttt{LatentDirichletAllocation} (batch variational inference, 20 iterations) with $K$ set to the in-scope class count of the dataset, fit on a \texttt{CountVectorizer} feature set with the standard English stop-word list, \texttt{min\_df}=2 and \texttt{max\_df}=0.95. The per-document cluster assignment is the argmax of the document--topic distribution; the per-topic top words from $\phi$ are saved as cluster descriptions but not used by any metric. \emph{Licences:} scikit-learn is BSD-3-Clause.

\paragraph{SBERT+$k$-means.} \texttt{sentence-transformers/all-mpnet-base-v2} sentence embeddings (768-dim), L2-normalised, fed to scikit-learn's $k$-means with \texttt{n\_init}=10 and a maximum of 300 iterations. L2-normalisation turns Euclidean $k$-means into cosine $k$-means on the unit sphere, the standard convention for SBERT embeddings. Embeddings are cached on disk so that subsequent methods using the same encoder (e.g.\ BERTopic) reuse them rather than recomputing. \emph{Licences:} \texttt{all-mpnet-base-v2} is Apache-2.0; the \texttt{sentence-transformers} library is Apache-2.0; scikit-learn is BSD-3-Clause.

\paragraph{BERTopic.} The standard SBERT $\to$ UMAP $\to$ HDBSCAN $\to$ class-based TF-IDF pipeline of \citet{grootendorst2022bertopic}, using the same \texttt{all-mpnet-base-v2} embeddings as SBERT+$k$-means (the cache is shared). UMAP is configured with \texttt{n\_neighbors}=15, \texttt{n\_components}=5, \texttt{min\_dist}=0.0, cosine metric, and an explicit \texttt{random\_state} for reproducibility; HDBSCAN with \texttt{min\_cluster\_size}=10 and the default EOM cluster-selection method; c-TF-IDF over a \texttt{CountVectorizer} with the standard English stop-word list. We report two configurations per dataset: a given-$k$ run with \texttt{nr\_topics} set to the in-scope class count, which uses BERTopic's post-hoc hierarchical merging step to reduce HDBSCAN's discovered topics to the requested number, and a discover-$k$ run with \texttt{nr\_topics=None} that keeps HDBSCAN's natural output. Per the noise-handling rule of \S\ref{sec:experiments}, we call \texttt{reduce\_outliers(strategy=`embeddings')} on the five datasets without a gold ``none'' class to force-assign every document, and skip it on CLINC and GoEmotions so that HDBSCAN's noise cluster ($-1$) is retained as BERTopic's native ``no cluster'' output and aligned to the gold $-1$ class. \emph{Licences:} BERTopic, UMAP, and HDBSCAN are all MIT-licensed.

\paragraph{LLM-embedding+$k$-means.} OpenAI \texttt{text-embedding-3-large} embeddings (3072-dim), obtained through the OpenAI Batch API (50\% discount, $\leq$24h SLA) per \S\ref{sec:experiments}, then fed to the same scikit-learn $k$-means clusterer as SBERT+$k$-means (\texttt{n\_init}=10, 300 iterations maximum, L2-normalised, $K$ set to the in-scope class count). Document inputs are truncated to the encoder's 8{,}192-token ceiling before embedding. The resulting vectors are cached on disk together with the actual token usage and USD, so the cost column reports what we paid rather than a re-estimate. This is the LLM-embedding pipeline reported by \citet{petukhova2024llm}, instantiated with a current frontier embedder. \emph{Licences:} OpenAI \texttt{text-embedding-3-large} is a proprietary commercial service used under the OpenAI API Terms of Use; scikit-learn is BSD-3-Clause.

\paragraph{ClusterLLM.} We run the triplet (``perspective'') variant of ClusterLLM \citep{zhang-etal-2023-clusterllm}, with given-$k$, from the authors' code with its default hyperparameters throughout (Instructor-large embeddings, \texttt{hkunlp/instructor-large}; 1{,}024 sampled triplets; the upstream \texttt{scripts/finetune.sh} fine-tuning settings). We make two changes. First, triplet judging is run with GPT-5-mini via the OpenAI Batch API. Second, we drive Instructor through \texttt{sentence-transformers}~5.x's \texttt{prompt=} API rather than the authors' \texttt{sentence-transformers}~2.2 wrapper, which no longer runs under current libraries; weights and instructions are unchanged. \emph{Licences:} the upstream ClusterLLM code is released without a LICENCE file and is vendored here under a research-reproduction rationale; \texttt{hkunlp/instructor-large} is Apache-2.0; GPT-5-mini is a proprietary OpenAI service used under the OpenAI API Terms of Use.

\paragraph{TopicGPT.} We run TopicGPT \citep{pham-etal-2024-topicgpt} from the authors' code at recursion depth~1, producing a flat taxonomy that matches our gold-label format. Topic generation and refinement run on the frontier tier (Claude Opus 4.7 via the Claude Code subscription); per-document assignment and the follow-up correction pass run on the cheap tier (GPT-5-mini via the OpenAI Batch API). For topic generation, we provide no seed topics, so the taxonomy is discovered from the corpus alone --- matching the fully-unsupervised, no-privileged-information setting we apply to every LLM baseline --- and we use the paper's recommended \texttt{early\_stop}=200. We add a defensive cap of 1{,}000 documents scanned for datasets where the model keeps discovering new topics and never triggers early stop (GoEmotions and StackExchange), as it would otherwise run through the whole corpus. \emph{Licences:} the upstream TopicGPT code is released without a LICENCE file and is vendored here under a research-reproduction rationale; Claude Opus 4.7 and GPT-5-mini are proprietary services used under the Anthropic and OpenAI API Terms of Use respectively.

\paragraph{Huang \& He.} We run \citet{huang-he-2024-text-clustering} from the authors' code with its default hyperparameters (label generation over chunks of $B{=}15$ documents; the generate~$\to$~merge~$\to$~classify pipeline with no outer iteration). We run their 0\%-seed configuration --- passing an empty \texttt{given\_labels} list into the verbatim upstream label-generation prompt --- rather than their 20\%-seeded headline, which injects gold-label names and would give the method privileged information no other baseline receives. Label generation and per-document classification run with GPT-5-mini via the OpenAI Batch API; the single label-merging call per dataset runs on the frontier tier (Claude Opus 4.7 via the Claude Code subscription). \emph{Licences:} the upstream Huang \& He code is released without a LICENCE file (the upstream GitHub API reports \texttt{"license": null}) and is vendored here under a research-reproduction rationale; Claude Opus 4.7 and GPT-5-mini are proprietary services used under the Anthropic and OpenAI API Terms of Use respectively.

\paragraph{Our method.} For all benchmark runs we use the default classification prompt, built directly from the finalised taxonomy as described in \S\ref{sec:method-classification}, and do not run the optional prompt-tuning step. The only labels available on the benchmark datasets are the gold partition itself, so tuning the classifier against them would leak the evaluation signal into the method under test. \emph{Licences:} our agentic-clustering implementation is released under the MIT License; the underlying Claude Opus 4.7 and GPT-5-mini models are proprietary services used under the Anthropic and OpenAI API Terms of Use respectively.

\section{Seed variance}
\label{sec:appendix-seeds}
\begin{table*}[t]
  \centering
  \small
  \begin{tabular}{l c cccc}
    \toprule
    \textbf{Dataset} & $k$ & ARI & NMI & ACC & ARI (in-scope) \\
    \midrule
    Banking77 & 81 {\scriptsize $\pm$ 4} & 0.708 {\scriptsize $\pm$ 0.048} & 0.883 {\scriptsize $\pm$ 0.015} & 0.810 {\scriptsize $\pm$ 0.042} & --- \\
    CLINC150 & 166 {\scriptsize $\pm$ 13} & 0.598 {\scriptsize $\pm$ 0.229} & 0.927 {\scriptsize $\pm$ 0.015} & 0.879 {\scriptsize $\pm$ 0.058} & 0.892 {\scriptsize $\pm$ 0.014} \\
    MASSIVE-Intent & 62 {\scriptsize $\pm$ 3} & 0.609 {\scriptsize $\pm$ 0.074} & 0.825 {\scriptsize $\pm$ 0.019} & 0.717 {\scriptsize $\pm$ 0.085} & --- \\
    MASSIVE-Domain & 20 {\scriptsize $\pm$ 2} & 0.648 {\scriptsize $\pm$ 0.026} & 0.776 {\scriptsize $\pm$ 0.014} & 0.750 {\scriptsize $\pm$ 0.031} & --- \\
    GoEmotions & 25 {\scriptsize $\pm$ 4} & 0.088 {\scriptsize $\pm$ 0.008} & 0.243 {\scriptsize $\pm$ 0.006} & 0.330 {\scriptsize $\pm$ 0.020} & 0.239 {\scriptsize $\pm$ 0.020} \\
    20 Newsgroups & 21 {\scriptsize $\pm$ 1} & 0.568 {\scriptsize $\pm$ 0.014} & 0.677 {\scriptsize $\pm$ 0.005} & 0.740 {\scriptsize $\pm$ 0.012} & --- \\
    StackExchange & 119 {\scriptsize $\pm$ 20} & 0.448 {\scriptsize $\pm$ 0.022} & 0.742 {\scriptsize $\pm$ 0.008} & 0.599 {\scriptsize $\pm$ 0.009} & --- \\
    \bottomrule
  \end{tabular}
  \caption{Mean $\pm$ standard deviation over 3 seeds in the discover-$k$ setting.
  \emph{ARI (in-scope)} recomputes ARI over documents whose gold label is not the
  ``none'' class, and is reported only for the two datasets that have one.}
  \label{tab:seed-variance}
\end{table*}

\begin{table*}[b]
  \centering
  \small
  \begin{tabular}{l ccc ccc ccc}
    \toprule
     & \multicolumn{3}{c}{\textbf{Full (discover-$k$)}} & \multicolumn{3}{c}{\textbf{$-$ refinement loop}} & \multicolumn{3}{c}{\textbf{$-$ task description}} \\
    \cmidrule(lr){2-4}\cmidrule(lr){5-7}\cmidrule(lr){8-10}
    \textbf{Dataset} & ARI & NMI & ACC & ARI & NMI & ACC & ARI & NMI & ACC \\
    \midrule
    Banking77 & 0.65 & 0.87 & 0.76 & 0.64 & 0.86 & 0.76 & 0.69 & 0.87 & 0.77 \\
    CLINC150 & 0.81 & 0.94 & 0.93 & 0.82 & 0.94 & 0.93 & 0.39 & 0.88 & 0.69 \\
    MASSIVE-Intent & 0.55 & 0.81 & 0.65 & 0.53 & 0.80 & 0.64 & 0.68 & 0.85 & 0.82 \\
    MASSIVE-Domain & 0.62 & 0.76 & 0.71 & 0.63 & 0.76 & 0.72 & 0.60 & 0.76 & 0.70 \\
    GoEmotions & 0.08 & 0.24 & 0.31 & 0.08 & 0.24 & 0.31 & 0.07 & 0.19 & 0.27 \\
    20 Newsgroups & 0.58 & 0.67 & 0.73 & 0.56 & 0.68 & 0.72 & 0.50 & 0.64 & 0.63 \\
    StackExchange & 0.46 & 0.75 & 0.61 & 0.45 & 0.74 & 0.59 & 0.45 & 0.74 & 0.61 \\
    \bottomrule
  \end{tabular}
  \caption{Component ablations of our method in the discover-$k$ setting ($k\pm20\%$). \emph{Full} is the complete method; \emph{$-$ refinement loop} classifies against the synthesizer's first proposed taxonomy, skipping the auditor, critic and investigator; \emph{$-$ task description} runs the full method without the user's task description.}
  \label{tab:ablations}
\end{table*}

\noindent The ranking does not depend on the seed. Running our method at three
seeds in the discover-$k$ setting, every individual run beats the strongest
discover-$k$ baseline on every dataset and on all three metrics --- 63 of 63
comparisons --- and the three-seed mean leads by at least 1.5 standard
deviations, on four datasets by more than seven
(Table~\ref{tab:seed-variance}). Each run resamples every stochastic stage:
proposer sampling, agent dispatch, orchestrator decisions, and the final
classification pass.

Variance is nonetheless strongly dataset-dependent rather than a single property
of the method: ARI standard deviations range from 0.008 on GoEmotions to 0.229 on
CLINC150. In the outlying case it has a specific source --- our method sometimes
partitions the ``none'' class. The refinement loop treats the out-of-scope pool
as an uncovered region of the corpus and carves coherent clusters from it --- on
CLINC150 one run produced \emph{general knowledge / trivia} and \emph{household
how-to}, assigning 711 of the 1,000 out-of-scope queries to in-scope clusters
where another run assigned only 109. Whether the loop does this is the largest
single source of run-to-run variation we observe, and it separates the metrics
exactly as described in the experimental setup: NMI tolerates the redistribution,
ARI and ACC penalise it. Restricted to in-scope documents all three metrics are
stable (CLINC150 ARI sd 0.014 and ACC sd 0.009, against 0.229 and 0.058 overall);
the same effect suppresses GoEmotions, whose in-scope ARI is 0.239 against 0.088.
We keep the ``none'' class in the headline numbers, but ARI and ACC on these two
datasets measure clustering and out-of-scope rejection jointly.

\section{Ablation results}
\label{sec:appendix-ablations}

\noindent This appendix gives the full results of the ablations discussed in \S\ref{sec:results} (``Ablations''): two component ablations, which remove part of the method (Table~\ref{tab:ablations}), and two design ablations, which vary how the proposers are run (Table~\ref{tab:design-ablations}). Both component ablations use the discover-$k$ configuration. The \emph{$-$ refinement loop} row reuses each run's first synthesizer output, recovered from the run workspace, and re-runs only the cheap-tier classification against it; no frontier calls are incurred. The \emph{$-$ task description} row is a full end-to-end run with \texttt{config.instructions} left empty --- the plugin's documented ``discover from the data alone'' mode --- and the dataset name removed from the orchestrator prompt, with the $k$-range and every other setting held identical to the main discover-$k$ run, so the task description is the only changed variable.

\paragraph{Design ablations.} The two design ablations each vary a single setting of the proposers --- how many are dispatched, and how many texts each one reads --- and observe the orchestrator's response (Table~\ref{tab:design-ablations}). All are discover-$k$ and single-seed, so we report the summed cluster-count deviation $\Sigma|\Delta k|$ (the summed absolute difference between the discovered and gold cluster counts across the seven datasets) rather than the partition metrics of Table~\ref{tab:results}.

\begin{table}[t]
  \centering
  \small
  \begin{tabular}{l ccc}
    \toprule
    \textbf{Front-end setting} & $\Sigma|\Delta k|$ & \makecell{mean\\investig.} & \makecell{mean\\dispatch} \\
    \midrule
    \multicolumn{4}{l}{\textit{Proposer count}} \\
    \midrule
    2--3 proposers & 50 & 5.3 & 16.4 \\
    4--5 proposers & 33 & 6.1 & 19.3 \\
    \textbf{6--7 proposers (default)} & \textbf{21} & \textbf{5.9} & 21.0 \\
    8--9 proposers & 58 & 5.3 & 21.3 \\
    \midrule
    \multicolumn{4}{l}{\textit{Sample size (per proposer)}} \\
    \midrule
    0.5$\times$ & 19 & 7.9 & 24.9 \\
    \textbf{1.0$\times$ (default)} & \textbf{21} & \textbf{5.9} & 21.0 \\
    2.0$\times$ & 25 & 4.3 & 18.7 \\
    \bottomrule
  \end{tabular}
  \caption{Design ablations (discover-$k$, aggregated across all benchmark
  datasets). The upper panel varies the number of proposers dispatched in
  parallel before synthesis; the lower panel varies how many texts each proposer
  samples, as a multiple of the default. All other settings are held at their
  production values. $\Sigma|\Delta k|$ is the summed absolute deviation of the
  discovered cluster count from gold; lower is better. \emph{Investig.} and
  \emph{dispatch} are the mean per-dataset counts of investigation rounds and
  total agent dispatches.}
  \label{tab:design-ablations}
\end{table}

\emph{Proposer count.} Varying the number of parallel proposers traces a clean optimum: $\Sigma|\Delta k|$ falls from 50 (2--3 proposers) through 33 (4--5) to 21 (6--7), then rises to 58 (8--9). Too few proposers miss low-frequency categories; too many flood the synthesizer with partially-disagreeing drafts.

\emph{Sample size.} Halving each proposer's sample (0.5$\times$) preserves final quality ($\Sigma|\Delta k|$ 19 vs.\ 21) but raises the investigation rounds the orchestrator needs to get there (mean 5.9 $\to$ 7.9). The post-synthesis refinement loop is what closes the gap: the halved front-end hands it a first-pass taxonomy that is far off ($\Sigma|\Delta k|$ 57, against 26 at the default sample), and the auditor, critic and investigator between them correct 38 points of that deviation, against 5 at the default. Doubling the sample (2.0$\times$) shows the same pattern: a first pass of 47 corrected by 22 points, finishing at 25 on fewer investigation rounds (4.3). This is the clearest evidence that the refinement loop is a working, adaptive component rather than a fixed pass.

\let\origunderscore\_
\renewcommand{\_}{\origunderscore\allowbreak}
\section{Qualitative analysis}
\label{sec:appendix-qualitative}

\noindent The metrics in Section~\ref{sec:results} summarise how well our clusters match the gold labels, but not what the clusters are or why they differ from gold. This appendix examines the method's output and behaviour directly. We first present a complete taxonomy and compare it with the gold labels (\S\ref{sec:qual-taxonomy}), then trace how the refinement loop changes the taxonomy, including cases where it gets it wrong (\S\ref{sec:qual-refinement}), how the method handles texts that sit on the boundary between clusters (\S\ref{sec:qual-boundaries}), and how it assigns queries to no cluster (\S\ref{sec:qual-oos}).

\subsection{A complete taxonomy (MASSIVE-Domain)}
\label{sec:qual-taxonomy}

To show what the method actually produces, Table~\ref{tab:qual-taxonomy} lists one complete final taxonomy: the 20 clusters our method found on MASSIVE-Domain in the discover-$k$ setting (seed 0). MASSIVE-Domain contains 2,974 voice-assistant utterances, each labelled with one of 18 gold domains. For each predicted cluster, Table~\ref{tab:qual-taxonomy} gives the description exactly as passed to the classifier and the gold domain most of its utterances belong to. We then compare the taxonomy with those gold domains, to show where and why the two disagree.

\begin{table*}[p]
  \centering
  \footnotesize
  \renewcommand{\arraystretch}{1.15}
  \begin{tabular}{@{}>{\raggedright\arraybackslash}p{2.5cm}p{9.9cm}>{\raggedright\arraybackslash}p{2.2cm}@{}}
    \toprule
    \textbf{Predicted cluster} & \textbf{Description} & \textbf{Majority gold domain} \\
    \midrule
    Email, messaging and contacts & Private, directed communication and contact-book lookups: composing, sending, replying to, checking, filtering, or setting notifications on email; placing or answering voice calls; dispatching direct 1-to-1 messages (SMS, chat) to a named person; and looking up personal contact information (phone numbers, contact groups, where a known contact lives). The connective tissue is a one-to-one, addressed, private communication surface, or the address book that supports it. Public posts belong in Social media; broad factual queries about celebrities or public figures belong in General knowledge. & email (0.93) \\
    Smart-home and device control & Commands that adjust connected household devices and on-device parameters: lights (on/off, colour, brightness), plugs and sockets, vacuum, coffee-maker, thermostat, HVAC, speaker or playback volume, and generic app-open or app-close commands. The interaction is a state-change on a connected physical device or a system-level control. Lighting commands require an explicit lighting device (light, lamp, bulb) --- a bare descriptor such as a colour, artist, or song title on its own is not a smart-home command and belongs to the cluster its content most naturally describes. & iot (0.77) \\
    Calendar and meetings & Scheduling, editing, cancelling, or querying calendar events, meetings, appointments, and other dated personal items (dentist, birthdays as dated events, meeting-anchored alerts). Anchored to a date/time slot, often with attendees or a location. Distinct from free-form reminder pings and from time/date lookups. & calendar (0.96) \\
    General knowledge and math & Open-domain factual, definitional, biographical, and arithmetic queries: `who is X', `define Y', `what is X plus Y', geographical facts, celebrity trivia, encyclopaedic lookups, speculative/philosophical questions, and generic `tell me about X' or `web search' style requests. Static information; time-sensitive current events belong in News, personal-contact directory lookups belong in Email/messaging, and weather/finance/cooking questions belong in their specialised clusters. & qa (0.75) \\
    Music playback & Requests to play, skip, replay, rate, or otherwise control songs, artists, albums, playlists, or genres; queries about the currently playing track; and stated music preferences. Includes short lyric or song-title fragments that read as truncated play commands. Playback of live radio stations, podcasts, or audiobooks belongs in the audio-streaming cluster instead, not here. & play (0.75) \\
    Small talk and assistant meta & Non-utility conversation with the assistant: greetings, jokes, personal disclosures about the user's day or mood, opinion prompts, and meta-comments about the assistant's own audio, wording, or behaviour. Also the natural home for conversational fragments that carry no specialised domain signal. Fragments that DO carry a clear functional signal --- a filter rule, a contact-directed request, a factual query, a time or list operation --- route to their functional cluster even when terse; c19$^\dagger$ is only for fragments with no such signal. & general (0.52) \\
    Local search and nearby places & Location-aware lookups grounded in the user's neighbourhood: nearby shops and stores, restaurants, food festivals, movie theatres and showtimes, `what's happening around me' queries, plus takeaway ordering and food-order status. The connective tissue is `find or transact with local businesses/events'. & recommendation (0.47) \\
    Weather & Weather queries and weather-conditioned decisions: forecasts, current conditions, temperature, precipitation, and the practical follow-on questions whose answers depend on the weather (clothing choices, driving-gear decisions, whether to bring sunglasses or an umbrella). Any question whose resolution requires a weather lookup belongs here, not in general knowledge or small talk. & weather (0.95) \\
    Radio, podcasts and audiobooks & Playback of non-music streaming or broadcast audio: tuning to live radio stations (FM/AM/internet), starting or resuming named podcasts and talk shows, and resuming or navigating audiobooks. Grouped as one coarse audio-streaming domain distinct from music because the primary object is a station or long-form spoken program rather than a chosen track. & play (0.98) \\
    \bottomrule
  \end{tabular}
  \caption{The complete final taxonomy produced by our method on MASSIVE-Domain (discover-$k$,
  seed 0), ordered from largest to smallest cluster and continued overleaf. Descriptions are
  reproduced verbatim as passed to the classifier. The last column gives the gold domain most of
  each cluster's utterances belong to, with its share.
  $^\dagger$This description refers to its own cluster by its internal identifier, which the
  classifier sees alongside the cluster's name.}
  \label{tab:qual-taxonomy}
\end{table*}
\begin{table*}[p]
  \centering
  \footnotesize
  \renewcommand{\arraystretch}{1.15}
  \begin{tabular}{@{}>{\raggedright\arraybackslash}p{2.5cm}p{9.9cm}>{\raggedright\arraybackslash}p{2.2cm}@{}}
    \toprule
    \textbf{Predicted cluster} & \textbf{Description} & \textbf{Majority gold domain} \\
    \midrule
    News and current events & Requests for time-sensitive current-events content: news headlines, named outlets, political and world-affairs updates, breaking-topic alerts, sports scores and schedules (upcoming or ongoing matches), and TV or entertainment recaps (`what happened on X show'). Distinct from static general-knowledge lookups: anything whose answer would change hour-to-hour or day-to-day belongs here. & news (0.78) \\
    Reminders and personal tasks & Setting, listing, or clearing `remind me to X' style task pings that fire at a future time --- chores, personal errands, one-off or recurring habit prompts. Distinct from Calendar because there is no dated event with attendees, and distinct from Lists because the output is a scheduled ping rather than an item added to a persistent named collection. & calendar (0.93) \\
    Travel and navigation & Getting-somewhere queries and transit planning: driving directions, live traffic conditions, travel-time and ETA lookups, and booking or querying train/plane tickets between cities. Merged as a single coarse transport cluster because ticket-booking and navigation both anchor to `moving between places' at this level of abstraction. & transport (0.92) \\
    Lists --- shopping and to-do & Creating, editing, reading, clearing, adding to, or removing items from user-managed lists: shopping, groceries, to-do, and other named collections. The interaction operates on a persistent list object. Time-triggered task pings belong in Reminders even when they concern the same underlying tasks. & lists (0.97) \\
    Time and date & Current clock or date queries, cross-city time comparisons, and time-zone conversions. Pure temporal-orientation lookups with no task attached. Alarms and calendar events live in their own clusters even when they involve a time. & datetime (0.96) \\
    Alarms & Setting, removing, listing, or querying wake-up alarms tied to a specific clock time or day. The interaction is with a time-triggered ringing/wake surface. Scheduled events with attendees or dates belong in Calendar, and free-form `remind me to X' task pings belong in Reminders. & alarm (0.98) \\
    Social media and public reviews & Posting to and reading from public social platforms (Facebook, Twitter, Instagram) --- status updates, tweets, feed scans, live-video queries, notifications --- plus consumer complaints and public reviews addressed to companies (tweeted complaints, review-site postings, brand-directed grievances). Grouped because all share a public-broadcast/one-to-many surface; private messaging belongs in Email/messaging. & social (0.98) \\
    Cooking and recipes & Recipe lookups, ingredient substitutions, cooking times and temperatures, technique questions, cookware comparisons, and other kitchen-domain knowledge. Kept separate from general knowledge because voice assistants route these to a dedicated recipe skill. Takeaway ordering and food-order status belong in Local search. & cooking (0.93) \\
    Finance --- stocks and currency & Stock-price lookups for specific tickers and currency exchange-rate queries. Tight market-data domain with a distinct financial backend, kept separate from general knowledge. & qa (0.94) \\
    Games & Requests for the assistant to play a game with the user, including named on-device games (chess, Temple Run, etc.). Small but distinct interactive-entertainment surface with its own game-engine backend, kept separate from small talk and from media playback. & play (0.97) \\
    Video, TV and movies & Playback and discovery of on-demand video content: launching streaming services (Netflix, YouTube, etc.) on a TV or console, playing named shows or videos, browsing trending lists, and `what should I watch' / `recommend me a movie' style discovery queries. The connective tissue is a video/TV-streaming backend surface. Distinct from audio-only streaming (radio, podcasts, audiobooks) because the primary object is a visual programme; distinct from local-cinema queries (which route on physical showtimes and theatre locations); distinct from movie trivia lookups about actors, awards, or filmography (which are static encyclopaedic questions). & recommendation (0.57) \\
    \bottomrule
  \end{tabular}

  \medskip
  \small Table~\ref{tab:qual-taxonomy} (continued).
\end{table*}

\paragraph{Subdivided gold domains.} Three gold domains are each divided across more than one predicted cluster. Utterances in the gold domain \emph{play} are divided between the predicted clusters \emph{Music playback} (``play nineties hip hop''), \emph{Radio, podcasts and audiobooks} (``play my favorite podcast from list'') and \emph{Games} (``olly play poker with me''). Utterances in the gold domain \emph{qa} are divided between the predicted clusters \emph{General knowledge and math} and \emph{Finance --- stocks and currency} (``how have megatel shares done last week''). Utterances in the gold domain \emph{calendar} are divided between the predicted clusters \emph{Calendar and meetings} and \emph{Reminders and personal tasks} (``set repeating reminder for every sunday''). Each of these predicted clusters draws almost entirely on a single gold domain, so these are defensible refinements, and they account for our method finding 20 clusters where the gold labels have 18.

\paragraph{Merged gold domains.} In three places, two gold domains end up in the same predicted cluster, and in each case the grouping is written into that cluster's description. The predicted cluster \emph{Smart-home and device control} covers ``speaker or playback volume'', so it contains both the gold domain \emph{iot} (``turn the lights to red color'') and most of the gold domain \emph{audio} (``raise volume'', ``mute for fifteen minutes''). The predicted cluster \emph{Music playback} covers ``queries about the currently playing track'', so it contains both the gold domain \emph{play} and most of the gold domain \emph{music} (``what song is on the radio''). The predicted cluster \emph{Local search and nearby places} explicitly includes ``takeaway ordering and food-order status'', so it contains both the gold domain \emph{recommendation} (``where can i go tonight'') and the gold domain \emph{takeaway} (``i need to order some food from papa john's''). The first two groupings are defensible; the third, which treats placing a food order as a kind of local search, is the least convincing.

\paragraph{Context-dependent gold labels.} The weakest predicted cluster, \emph{Small talk and assistant meta}, collects utterances from many gold domains. In these cases the gold label reflects which device would handle the request rather than anything in the wording: ``i need a cup of coffee'' is labelled with the gold domain \emph{iot} (a smart coffee machine), and ``can you please stop speaking'' with the gold domain \emph{audio}. The gold domain \emph{general} is itself a catch-all, and its utterances are split between this predicted cluster and the predicted cluster \emph{General knowledge and math}.

\subsection{Refinement in action}
\label{sec:qual-refinement}

We trace four changes the refinement loop made to a taxonomy: two where refinement improved agreement with the gold labels, and two where it fell short. Each is described in five steps: the initial definition, how the problem was detected, the investigation, the change, and the outcome measured against the gold labels. All examples are from the discover-$k$ setting (seed 0).

\begin{qualexample}{Example A: Rewriting a misread definition (Banking77)}
\emph{Banking77 contains 3,080 customer queries to a digital bank, each labelled with one of 77 fine-grained gold categories. The synthesizer's first taxonomy had 81 clusters, and refinement ended at 83.}

\smallskip
\textbf{(1) Initial definition.} One proposer's sample contained ``I didn't receive the right amount of cash back''. On its own the phrase is ambiguous: it can mean a loyalty-cashback reward that was short, or cash the customer didn't get. The synthesizer took the loyalty reading and created the predicted cluster \emph{Wrong cashback amount received}: ``Customer received an incorrect amount of cashback and wants it corrected.'' In fact the text is about an ATM, with the gold category \emph{wrong\_amount\_of\_cash\_received}.

\textbf{(2) Detection.} On unseen texts, the auditor could fit ATM complaints into this cluster only at low confidence (mean 2.4/5), which flagged it as weak.

\textbf{(3) Investigation.} The investigator ran TF-IDF searches for both readings. It found at least 15 ATM wrong-amount texts, several using ``cash back'' to mean \emph{return my missing cash}: ``I have got out the wrong amount from an ATM. Please tell me the process to get the cash back\ldots'' and ``How can I request cash back? The ATM just gave me the wrong amount\ldots''. It found no texts at all about loyalty cashback. Rather than adding a new cluster alongside this one, it recommended rewriting this cluster's definition.

\textbf{(4) Revised definition.} \emph{Wrong amount of cash received from ATM}: ``Customer withdrew cash at an ATM and the machine dispensed less (or more) than the amount requested\ldots{} Customers often phrase this as wanting `cash back', meaning the missing cash back to them, not a loyalty-reward cashback product. Distinct from a declined withdrawal, a fee dispute\ldots, an exchange-rate dispute\ldots, or a pending withdrawal.''

\textbf{(5) Outcome.} Audit confidence rose to 4.3--4.75. The final predicted cluster has 48 texts, 79\% in the gold category \emph{wrong\_amount\_of\_cash\_received}.
\end{qualexample}

\begin{qualexample}{Example B: Adding a missing cluster (Banking77)}
\textbf{(1) Initial taxonomy.} The synthesizer created three predicted PIN clusters: \emph{Reset forgotten PIN / passcode / password}, \emph{PIN blocked after too many wrong attempts}, and \emph{Set up a new PIN} (``\ldots setting up a PIN on a new card''). None covered a customer who has a working PIN and wants to change it.

\textbf{(2) Detection.} The auditor placed unseen texts such as ``I would like to change my card PIN'' and ``I would like to change my PIN but I am not currently in the country'' in the nearest PIN cluster at confidence 2--3, with notes like ``no dedicated change-PIN cluster; [\emph{Set up a new PIN}] is `set up on new card', [\emph{Reset forgotten PIN}] is `forgot' '', and the critic separately flagged the gap.

\textbf{(3) Investigation.} Searching the corpus turned up 40 texts with this need, none about a forgotten, blocked or first-time PIN. They included recurring variants such as changing the PIN at an ATM and changing it while abroad. That is far too many to treat as stray texts, and the $k$-range had room for another cluster, so the investigator recommended adding one.

\textbf{(4) New definition.} \emph{Change existing PIN}: ``Customer already has a working PIN for their card and wants to change it to a different value. Includes questions about where the change can be done (ATM, in-app, at a branch), doing it while travelling or abroad\ldots{} Does not cover forgotten or unknown PINs\ldots, PINs blocked by too many wrong entries\ldots, or activating the initial PIN on a newly issued card\ldots''

\textbf{(5) Outcome.} In every later audit, change-PIN texts (14 in all) went to the new cluster at confidence 5. The final predicted cluster has 34 texts, all in the gold category \emph{change\_pin}, covering 34 of the corpus's 40 \emph{change\_pin} texts.
\end{qualexample}

\begin{qualexample}{Example C: Merging clusters that were small, not spurious (StackExchange)}
\emph{StackExchange contains 4,156 question titles from 121 Stack Exchange sites, with the site as the gold label.}

\smallskip
\textbf{(1) Initial taxonomy.} The synthesizer created separate predicted clusters for Ethereum, EOSIO and Tezos smart-contract development, matching three gold sites.

\textbf{(2) Detection.} Across several audits, including ones that deliberately sampled likely members of each cluster, the EOSIO and Tezos clusters received only 1--4 texts each and were flagged as persistently thin.

\textbf{(3) Investigation.} The investigator argued that all three serve the same user (developers writing on-chain code) and that ``separating them by chain produces persistently thin clusters at 4,156-text scale''. It recommended merging them.

\textbf{(4) Revised taxonomy.} A single predicted cluster, \emph{Smart-contract platform development}.

\textbf{(5) Outcome.} The merged cluster is 50\% \emph{ethereum}, 22\% \emph{tezos} and 20\% \emph{eosio}, so three gold sites were fused into one. The two smaller clusters were real but small: the gold sites \emph{eosio} and \emph{tezos} have only 19 and 15 titles in the corpus, and fewer than half of them name their chain (``How to calculate the txHash?'' could come from any blockchain site). Titles that did name the chain had been assigned correctly; the loop took the scarcity of such titles as evidence that the clusters were not real.
\end{qualexample}

\begin{qualexample}{Example D: A catch-all cluster mirroring catch-all gold labels (20 Newsgroups)}
\emph{20 Newsgroups contains 18,331 Usenet posts, with the newsgroup each post appeared in (one of 20) as the gold label.}

\smallskip
\textbf{(1) Initial definition.} The synthesizer created a predicted US-politics cluster whose description routed to it ``texts that don't fit the more specific guns/Waco/Mideast/religion buckets'', so it was defined by what it was not.

\textbf{(2) Detection.} The critic flagged it as a catch-all: the posts assigned to it ranged from rules of civil procedure to Iranian population growth.

\textbf{(3) Investigation.} Reading the 18 posts associated with the cluster, the investigator found three threads, with 3--8 examples each: US law and social-policy statistics; gender, sexuality and reproductive rights; and US foreign policy and government skepticism. It recommended a three-way split.

\textbf{(4) Reversal.} In the next audit, most texts targeted at the new foreign-policy cluster landed in other clusters, and the auditor reported that it could not tell the foreign-policy and US-law clusters apart. A second investigation merged them back into one cluster, again defined partly by exclusion, as US political commentary ``that isn't already carved off into a more specific subject cluster''. The gender cluster survived.

\textbf{(5) Outcome.} The resulting predicted cluster, \emph{US politics, law, and policy commentary}, is the second largest in the final taxonomy (1,314 posts), and only 33\% of its posts come from the gold newsgroup \emph{talk.politics.misc}. Two features of the gold labels contribute. First, a newsgroup records where a post appeared, not what it is about. Threads drift, so political arguments turn up across the corpus, and this cluster collects them: posts from the gold newsgroup \emph{sci.crypt} arguing about encryption law (``nobody's ever going to make me respect a law that violates my freedom of speech''), posts from \emph{sci.space} debating ``a right to a dark sky'', and posts from \emph{alt.atheism} arguing about Gulf War casualties. Second, the gold newsgroup \emph{talk.politics.misc} is itself a catch-all: its posts are split between this predicted cluster (57\%), \emph{Gender, sexuality, and reproductive-rights debate} (19\%) and \emph{Waco / Branch Davidian standoff} (11\%). The loop correctly identified its own cluster as a catch-all, but the material it was trying to divide is political argument with no single topic in the gold labels either, so its split did not hold.
\end{qualexample}

The two failures arise where the gold labels depend on information the text does not fully contain: in C, which site a question was posted to, when fewer than half the titles name their blockchain; in D, which newsgroup a post appeared in, when its thread has drifted from the group's topic. In C the loop also drew the wrong inference, taking a scarcity of identifiable titles as evidence that the clusters were not real.

\subsection{Ambiguous boundaries}
\label{sec:qual-boundaries}

Some texts fall between two clusters. We show how the method handles such boundaries at two stages: when the taxonomy is built, where the investigator decides whether two neighbouring clusters should stay separate, and when texts are classified, where the classifier must place texts close to the boundary between two descriptions.

\begin{qualexample}{Example E: A boundary correctly kept during discovery (MASSIVE-Intent)}
\emph{MASSIVE-Intent labels the same utterances as MASSIVE-Domain with 60 fine-grained gold intents. Among them, asking about current traffic (gold intent transport\_traffic) is a separate intent from asking for directions or travel times (gold intent transport\_query).}

\smallskip
The synthesizer's initial taxonomy contained two neighbouring predicted clusters, \emph{navigation\_directions\_or\_eta} and \emph{traffic\_conditions\_query}. In a targeted audit, only 2 of 5 utterances sampled as likely members of the traffic cluster were assigned to it, and the investigator was asked whether the traffic cluster should be merged into the navigation cluster or removed. Searching the corpus, it found at least 16 utterances asking about the state of traffic itself rather than for a route or arrival time (``what is the traffic like now'', ``is there a traffic jam on my route'', ``traffic report''). It rejected the merge, recording that the low audit count reflected the auditor confusing the two clusters rather than a shortage of texts, and recommended sharpening both descriptions so that each excludes the other:
\begin{itemize}
  \setlength{\itemsep}{1pt}
  \item \emph{navigation\_directions\_or\_eta}: ``Ask for routing to a destination: turn-by-turn directions, an ETA, or a travel-duration estimate. Focused on how to get somewhere --- not on the current condition of the road.''
  \item \emph{traffic\_conditions\_query}: ``Ask for a status report on current road-traffic conditions, congestion, jams, or accidents on a road or route. Focused on reporting the state of the road --- not on requesting directions or an ETA to a destination.''
\end{itemize}
This was the right decision. In the final classification, 14 of the 15 utterances in the gold intent \emph{transport\_traffic} fall in the predicted cluster \emph{traffic\_conditions\_query}, which contains almost nothing else.
\end{qualexample}

\begin{qualexample}{Example F: A boundary stated as a principle at classification (MASSIVE-Domain)}
Some voice-assistant requests depend on the weather without mentioning it: ``do i need a jacket'', ``should i bring an umbrella tomorrow''. All six proposers placed such questions with weather forecasts, and the synthesizer wrote this into the description of the predicted cluster \emph{Weather} as a general principle rather than a list of cases: it covers ``the practical follow-on questions whose answers depend on the weather (clothing choices, driving-gear decisions, whether to bring sunglasses or an umbrella). Any question whose resolution requires a weather lookup belongs here, not in general knowledge or small talk.''

\smallskip
The classifier applies the principle as intended. Of the 18 clothing or umbrella questions in the gold domain \emph{weather}, 17 are assigned to \emph{Weather}, with reasoning such as ``This is a weather-forecast-based clothing decision'' for ``will i need to wear a jacket next wednesday''. Clothing questions that do not depend on the weather stay out: ``what kinds of sunglasses are considered aviators'' and ``does kim kardashian wear converse'', both in the gold domain \emph{qa}, are assigned to \emph{General knowledge and math}, as ``a factual definitional question'' and ``celebrity\ldots{} trivia'' respectively. The one miss is ``should i wear sandals or shoes with socks'' (gold domain \emph{weather}), which the classifier read as ``a casual style/clothing-advice question'' and assigned to \emph{Small talk and assistant meta}. The text gives no hint that the answer depends on the weather, so the gold label is arguably the less obvious reading.
\end{qualexample}

\subsection{Correct and incorrect out-of-scope assignments}
\label{sec:qual-oos}

Some texts belong to no category. Our method allows for such texts at both stages. During discovery, proposers may leave texts unplaced and the auditor may assign ``none'' (\S\ref{sec:method-auditor}); for datasets that contain such texts, the orchestrator is also told that some texts will fit no cluster, so they are left unclustered rather than forced into one. At classification, the taxonomy passed to the classifier includes, alongside the discovered clusters, an \emph{Out of scope} category, described as applying when ``the text is genuinely outside the taxonomy, not just a poor fit.''

\begin{qualexample}{Example G: Out-of-scope queries (CLINC150)}
\emph{CLINC150 contains 5,500 queries to a virtual assistant: 4,500 across 150 gold intents (e.g.\ banking, travel, small talk) and 1,000 gold out-of-scope queries, many of them close to an in-scope intent.}

The classifier assigns 891 of the 1,000 out-of-scope queries to \emph{Out of scope}, and 96 of the 4,500 in-scope queries. In the correct cases the request falls outside every intent, as the classifier's reasoning states: ``can you group my photos by location'' is ``not covered by any banking, productivity, travel, or other listed categories.'' The errors have three recurring causes.

\textbf{Over-extension of a neighbouring cluster.} Some out-of-scope queries are assigned to a predicted cluster whose description does not cover them. ``what is the cashback rate for my credit card'' is assigned to \emph{rewards\_balance}, described as asking ``the current balance of rewards points / cashback'', although a rate is not a balance; ``restart my newspaper delivery for monday'' is assigned to \emph{order}, which the classifier itself describes only as the ``closest'' fit.

\textbf{Descriptions narrower than the gold intent.} ``how can i unfreeze my bank account'' belongs to the gold intent \emph{account\_blocked}, but the corresponding predicted cluster is described as asking ``why their bank account is on hold, frozen, or blocked''. The classifier noted that the taxonomy covers ``only\ldots{} asking why an account is blocked'' and assigned \emph{Out of scope}. The classifier applied the description correctly; the error lies in the description.

\textbf{Literal readings of indirect requests.} ``sorry i missed that'' (gold intent \emph{repeat}) is a request for the assistant to repeat itself, but the classifier read it as ``an apology'' and assigned \emph{Out of scope}.

A further set of disagreements reflects errors in the gold labels rather than the method. ``give me the weather forecast for today'' is labelled out-of-scope, although CLINC150 has a weather intent, to which the classifier assigns it. Conversely, five queries of the form ``i would like to change the name on my credit card account'' are labelled with the gold intent \emph{pin\_change}, and three concerning a lost school report card (``my report card was lost help!'') with the banking intent \emph{report\_lost\_card}; in each case \emph{Out of scope} is the more accurate assignment.
\end{qualexample}

\end{document}